\documentclass[journal]{IEEEtran}

\UseRawInputEncoding

\usepackage[table,xcdraw]{xcolor}
\usepackage{soul,framed} 

\colorlet{shadecolor}{yellow}
\usepackage[pdftex]{graphicx}
\graphicspath{{../pdf/}{../jpeg/}}
\DeclareGraphicsExtensions{.pdf,.jpeg,.png}

\usepackage[cmex10]{amsmath}
\usepackage{array}
\usepackage{mdwmath}
\usepackage{mdwtab}
\usepackage{eqparbox}
\usepackage{url}
\usepackage{amsmath}

\usepackage{adjustbox}
\usepackage{tabularx}
\usepackage{multirow}
\usepackage{rotating}
\usepackage{subcaption}    
\usepackage{algorithm, algpseudocode}
\usepackage{changepage} 

\usepackage{hyperref}
\usepackage{graphicx}

\newcommand{\orcid}[1]{\hspace{1mm}\href{https://orcid.org/#1}{\includegraphics[scale=0.5]{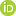}}\hspace{1mm}}

\begin{document}
\bstctlcite{IEEEexample:BSTcontrol}
\title{Segmentation of Drone Collision Hazards in Airborne RADAR Point Clouds Using PointNet}
\author{Hector Arroyo~\orcid{0009-0005-8133-8015},
    Paul Keir~\orcid{0000-0002-4781-9377},
    Dylan Angus~\orcid{0009-0003-1379-9499},
    Santiago Matalonga~\orcid{0000-0001-5429-2449},
    Svetlozar Georgiev~\orcid{0009-0007-6785-5694},\\
    Mehdi Goli~\orcid{0000-0002-3520-9598},
    Gerard Dooly~\orcid{0000-0002-7589-1384},
    James Riordan~\orcid{0000-0001-6516-9446}

}

\maketitle

\begin{abstract}
The integration of unmanned aerial vehicles (UAVs) into shared airspace for beyond visual line of sight (BVLOS) operations presents significant challenges but holds transformative potential for sectors like transportation, construction, energy and defense. A critical prerequisite for this integration is equipping UAVs with enhanced situational awareness to ensure safe operations. Current approaches mainly target single object detection or classification, or simpler sensing outputs that offer limited perceptual understanding and lack the rapid end-to-end processing needed to convert sensor data into safety-critical insights. In contrast, our study leverages radar technology for novel end-to-end semantic segmentation of aerial point clouds to simultaneously identify multiple collision hazards. By adapting and optimizing the PointNet architecture and integrating aerial domain insights, our framework distinguishes five distinct classes: mobile drones (DJI M300 and DJI Mini) and airplanes (Ikarus C42), and static returns (ground and infrastructure) which results in enhanced situational awareness for UAVs. To our knowledge, this is the first approach addressing simultaneous identification of multiple collision threats in an aerial setting, achieving a robust 94\% accuracy. This work highlights the potential of radar technology to advance situational awareness in UAVs, facilitating safe and efficient BVLOS operations.
\end{abstract}


%
\IEEEpeerreviewmaketitle

\section{Introduction}

\IEEEPARstart{T}ECHNOLOGICAL advances in miniaturization, autonomy, flight duration and payload capabilities have increased the possibilities of unmanned aerial vehicles (UAVs) to enable disruption in a wide range of industries, with an estimated three-fold growth in the next decade to 71 billion USD \cite{Https://bisresearch.com/industry-report/unmanned-aerial-vehicle-market.html}. The increased flight demand is expected to transform the way the airspace is operated: from flights primarily controlled by pilot input and cooperative protocols to unmanned beyond visual line of sight (BVLOS) flights in shared airspace. To allow safe and efficient navigation in this new environment, UAVs will require a higher level of autonomy to accompany any cooperative procedures. 

Widely researched sense and detect technologies are a first step towards increasing the situational awareness of aerial vehicles in the absence of the pilot’s physical see and avoid capabilities. Regulators \cite{EASA2021EASAApplications}, \cite{CAA2020BeyondTerminology}, \cite{FAA2022UnmannedUTM} are rapidly developing guidance for urban air mobility (UAM) operation \cite{Cohen2021UrbanChallenges} and are starting to break down the perception cycle, highlighting the objects of interest that the system should be aware of \cite{CAA2020DetectAirspace}. AI, specifically deep learning algorithms, have proven a powerful tool for processing large volumes of sensorial data and turning it into valuable insights. This has significant applications in air-to-air scenarios, enabling end-to-end data processing and analysis \cite{CAA2023EASAApplications}, \cite{Abdalla2020MachineSolutions}. 

Alongside advancements in perceptual understanding in other autonomous industries like automotive, there is a growing need for aerial systems to evolve and tackle more complex scenarios. These systems must be capable of simultaneously detecting and classifying multiple collision hazards in real-time as the scene unfolds, providing valuable information for safety critical manoeuvres like collision avoidance. 

Research efforts with a variety of sensors and setups have been made to advance the field of sense and detect. The most common setup is ground-to-air with a focus on the detection and classification of a single object. Little work has been done to understand the development constraints of air-to-air sense and detect systems and extend it towards end-to-end multi-object recognition as a first step to unlocking reliable and safe BVLOS flights. In this work we focus on improving the situational awareness of UAVs by enabling multi-object point cloud segmentation of collision hazards. This is a crucial task given the popularity of point clouds across different sensors including radar, lidar or ToF cameras; and its ease of integration with other common technologies such as RGB data. This enables onboard systems to recognize and classify objects, leading to informed decisions based on scene characteristics.

Furthermore, the sense-detect and avoid cycle, which holds a safety-critical role, is typically executed by power-constrained aerial vehicles. These vehicles operate in environments of limited bandwidth, latency, and uncertain network stability, privacy, or security. Optimizing this cycle is essential to rapidly and reliably translate sensor data into actionable insights, all while conserving energy and maintaining performance. This consideration gains significance in the context of designing solutions for aerial use cases, urging special focus on network optimization. By enhancing data flow efficiency from sensors to analysis, successful execution of the sense-detect and avoid cycle can be achieved, enabling informed decisions in challenging aerial scenarios.

This research presents a pioneering use case that leverages airborne radar technology for the semantic segmentation of point clouds. The algorithm demonstrates a distinctive capability in distinguishing between five unique classes of objects commonly expected in complex aerial scenes: large drones (DJI M300 RTK), small drones (DJI Mini), airplanes (Ikarus C42), ground returns, and man-made infrastructure. This marks an initial step towards equipping an aerial autonomous UAV with enhanced situational awareness in regards to collision hazards that can inform and improve decision-making processes for safe and efficient navigation. The main contributions of this paper we:

\par
\begin{enumerate}
    \item Advance the level of perceptual understanding for UAVs. By harnessing radar point clouds, we have achieved the effective segmentation of five prevalent classes within the aerial domain; as part of this we also
    \item Propose a workflow to evaluate the overall system performance in an aerial use case. It includes sensor performance characterization in the aerial paradigm, centered in radar and extensible to other sensing techniques, mission planning, dataset acquisition and data labelling of aerial point clouds; and
    \item Carry out a network optimization attending to (1) the higher-level building blocks of the network and (2) the lower-level network operators, and prove its superiority against a simpler machine learning counterpart for the task of collision hazard identification.
\end{enumerate}

The rest of the paper is organised as follows. Section \ref{sec:bg} introduces the related work on UAV sense and detect and situational awareness with different sensors and setups; Section \ref{sec:oa} describes our approach, including the scenario, system architecture and the development of the data pipeline, labelling, performance characterization and training of the network; Section \ref{sec:res} covers the experiments and results and finally Section \ref{sec:conc} presents the conclusions.

\section{BACKGROUND}
\label{sec:bg}

Situational awareness through collision hazard identification has become a key objective to achieving fully functional autonomous vehicles (AVs). In an aerial context, this challenge spans a full three-dimensional realm, with each dimension holding equal importance, as targets can potentially appear anywhere within this 3D space. This presents a departure from other paradigms such as automotive with a predominantly two-dimensional context where activity revolves around the road's horizontal plane.
Furthermore, aerial scenes typically exhibit low temporal variability, especially at higher altitudes, where encounters with other aerial objects are infrequent. Only at lower flying altitudes where ground and infrastructure data is gathered do these updates build up quickly. This contrasts with the automotive domain, where vehicle's ego-motion and high density of road actors drive a higher variability between consecutive updates.
The subsequent sections delve into this challenge, exploring sensor technologies, data processing methods and network selection and optimization strategies.

\subsection {Sensing Technologies}

Research related with UAVs has mainly been focused on air-to-ground or ground-to-air sense and detect scenarios with applications typically in air surveillance. The sensor is usually ground-based and scans the air in search of targets, usually UAVs or other small objects such as birds.  Lately, an increase in air-to-air sense and avoid within UAVs is taking place as a result of the increased capabilities of UAVs related to payload and flight time and continuous miniaturization of the sensors which allow their integration to a host drone.
Processing of sensor data has previously been carried out with approaches including custom algorithms, statistical methods (ML), and simple visual inspection of the returns. Deep learning methods for processing of the data have only recently started gaining traction as the field proves its capabilities. A summary of relevant work and our distinctive contributions is presented in the subsections below and summarised in Table 1.

\subsubsection {RGB sensor}

Systems that operate in the visible spectrum are the most explored arguably due to the maturity of the technology and its closeness to human’s  perception of the world. The capabilities of deep learning algorithms to discretise between drones and birds in ground-to-air UAV video images have proven succesful \cite{ArneSchumann2017DeepDetection} using background extraction of deep learning (DL) proposal methods. First steps towards increasing the situational awareness in an aerial context are taken using simulated video data of objects that a UAV would likely find during its operation \cite{Navarro2019SenseNetworks}, with a focus on network efficiency by using a combination of a convolutional neural network and a recurrent neural network CNN +RNN (CRNN) where the system uses an object detector in the first frame and then a custom recurrent neural network RNN (R3) in the remaining frames. An object detector is also used in other real air-to-air studies \cite{Leong2021Vision-BasedUAVs}, and applied to around 200 images of an UAV with a clear background that are segmented and cropped to extract the desired objects. These objects are then fused with other background images to increase the dataset size together with data augmentation with overall positive results. However, these results are focused on drone detection alone. Convolution neural networks {CNN} focusing on obstacle detection, as well as a reinforcement learning (RL) module for reactive obstacle avoidance for a UAV, are also explored \cite{Ma2018AObstacles} as a generic framework. On the detection side, the module uses a saliency detection algorithm leveraging deep CNNs to extract monocular visual cues. These studies above show that the building blocks of deep learning architectures can be helpful in an aerial context, however the studies focus mostly on synthetically generated data and not sequences of real world scenes.

Recent studies have started leveraging the capabilities of RGB data to perform identification of targets in an air-to-air setting. From developing an end to end deep learning pipeline to compare the performance of different object detection pipelines \cite{Zhao2022Vision-BasedTracking}, to exploring image segmentation techniques to identify aerial targets \cite{Minaeian2018EffectiveCamera}, both on UAV image acquired data. However, none of the studies tackle classification or semantic sementation to gain insights into the target's nature.

\subsubsection {Lidar}

Lidar is a remote sensing technology that uses laser pulses to measure distances and generate precise, three-dimensional maps. Accurate depth estimation and good resolution at range capabilities are native to lidar, at the expense of shorter range and lower resistance to adverse weather conditions. Lidar’s accurate returns for obstacle warning and avoidance have been used to classify ground and aerial, as well as static and moving objects \cite{Ramasamy2016LIDARSense-and-avoid}. Static vs dynamic object detection has also been explored \cite{Wang2018RobustEnvironments} with a heuristic method for object modelling that is subsequently used as an object tracker and velocity estimator, not native to lidar. However, no current studies approach sense and detect of UAVs from a lidar and deep learning point of view. A first attempt at presenting the challenges of lidar-based UAV to UAV sense and detect \cite{Riordan2021LiDARAvoid} is carried out  by leveraging a digital twin of the Port of Hamburg augmented with hazards such as birds, drones, helicopters and low-flying aircrafts. A raytracing-based lidar simulation is used to test sensing scenarios of objects located at different ranges. Specific examples at different ranges prove that sensing is highly dependent on the objects’ sizes and positions both in range and with respect to lidar’s FOV, with optimum detections happening at closer ranges and centred positions. Other works have explored the development of a UAV  algorithm with airborne LiDAR and an enhanced DBSCAN method for obstacle recognition and intrusion detection \cite{Miao2021AirborneEvaluation}, however it also lacks classification capabilities of targets. 

\subsubsection {Radar}

Radar technology operates by emitting radio waves and analyzing their reflections. It uses a transmitter to generate electromagnetic waves and a receiver to capture and process the reflected signals. Radar systems analyze the time it takes for the signals to return and the characteristics of the received signals to provide information about the object's distance, size, and velocity. Advanced signal processing techniques are employed to enhance detection and distinguish objects in noisy environments. In radar systems, a signal-to-noise ratio (SNR) threshold is used to filter out weaker signals. The SNR represents the ratio of the signal power to the noise power and is used to filter useful data. Signals above the SNR threshold, often set at the 3dB cutoff where the signal power drops to half its maximum, are retained for processing. These processed signals are used to generate a wide range of output formats, including radar spectrograms or three-dimensional (3D) point clouds. 

In addition to the spatial distribution of the returns, the radar cross section of each detection shows the amount of radar energy reflected back, which varies with the position of the return and scattering properties of the target. When processed with other features such as the location or velocity, it can be used to determine the size or the class of the target. Radar’s native velocity (doppler) information is particularly useful in an air to air context, as it enables, combined with location, a first step towards trajectory estimation of moving objects and safer navigation of the airspace.

\begin{center} 
\begin{table*}[t]
  \centering
  \caption{Summary of relevant work and our unique contribution}
  \begin{adjustbox}{width=7in,center}
  \renewcommand{\arraystretch}{1.5}
  \begin{tabular}{|p{1cm}|p{1.5cm}|p{2cm}|p{3cm}|p{3cm}|p{3cm}|p{2cm}|p{2cm}|p{3.5cm}|p{2.5cm}|}\hline
     & & \multicolumn{2}{|c|}{\textbf{\fontsize{10}{12}\selectfont{Data}}} & \multicolumn{3}{|c|}{\textbf{\fontsize{10}{12}\selectfont{Processing}}} & \multicolumn{3}{|c|}{\textbf{\fontsize{10}{12}\selectfont{Perceptual Understanding}}} \\\cline{3-10}
     & & \textbf{Synthetic/} & \textbf{Real} & \textbf{Visual/Custom} & \textbf{Statistical/ML} & \textbf{End to end DL} & \textbf{Sense} & \textbf{Single object detection / classification}  & \textbf{Multi-object recognition} \\\hline
     \multirow{3}{*}{\rotatebox{90}{\begin{tabular}{@{}c@{}}\textbf{\fontsize{12}{14}\selectfont{Ground}}\\ \textbf{\fontsize{12}{14}\selectfont{to air}}\end{tabular}}} 
     & \textbf{Camera} &{} & [9] &{} &{} & [9] &{} & [9] &{} \\ \cline{2-10}
     &\textbf{Lidar}  & & [16] & [16] & & & [16] & & \\ \cline{2-10}
     &\textbf{Radar}  & & [19],[20],[21],[22],[23],[24],
     [25],[26],[27],[28],[29],[30] & [19],[20],[21],[26],[27],[28],
     [29],[30] &[22],[23] &[24],[25] & [19] - [23], [26] - [30]& [24], [25]& \\\hline
 \end{tabular}
 \end{adjustbox}
\end{table*}
\vspace*{-1cm}
\begin{table*}[t]
  \centering
  \begin{adjustbox}{width=7in,center}
  \renewcommand{\arraystretch}{1.5}
  \begin{tabular}{!{\vrule width 3pt}p{1cm}|p{1.5cm}|p{2cm}|p{3cm}|p{3cm}|p{3cm}|p{2cm}|p{2cm}|p{3.5cm}|p{2.5cm}!{\vrule width 3pt}}\noalign{\hrule height 3pt}
    \multirow{3}{*}{\rotatebox{90}{\begin{tabular}{@{}c@{}}\textbf{\fontsize{12}{14}\selectfont{Air to}}\\ \textbf{\fontsize{12}{14}\selectfont{Air}}\end{tabular}}} 
    & \textbf{Camera} &[10],[11],[12] & [13],[14]& & [14]&[10],[11],[12],[13] &[11],[12] & [10],[13],[14] & \\ \cline{2-10}
     &\textbf{Lidar}  &[15],[17] &[18] &[15],[17] &[18] & &[15],[18] &[17],[18] & \\ \cline{2-10}
     &\textbf{Radar}  & [38]&[31],[32],[33],[34],[35],[36],
     [37], \textbf{[This work]}&[31],[33]& [32] &[34],[35],[36],[37],
     [38],\textbf{[This work]} &[31], [32], [33]&[34],[35],[36],[37],[38]&\textbf{[This work]} \\\noalign{\hrule height 3pt}
 \end{tabular}
 \end{adjustbox}
\end{table*}
\end{center}

Unlike light-based sensors, radio waves utilized by radar exhibit weak absorption in the air. Peaks of absorption of electromagnetic waves by water molecules occurs at bands between 22.2 GHz and 23.8 GHz, and around 60 GHz, with additional closely spaced resonances over the 50 to 70 GHz band. Radar sensors, such as the one employed in this study, often avoid the absorption peaks rendering an enhanced performance resilience against adverse weather phenomena like rain, fog, and snow. Additionally, radar remains unaffected by poor lighting conditions. Moreover, radar's built-in capability to detect objects at relatively long ranges is a crucial parameter for this application. These distinct features position radar as a likely fundamental component in future sense, detect, and classify systems when compared to other active and passive sensors.

 Characteristics of a target’s surface can be captured using radar polarimetry. This has been leveraged for long-range (up to 1.5km) sensing and classification of drones and birds on a ground-to-air setup and a classification pipeline based on K-Nearest Neighbors (KNN) \cite{TorvikClassificationPolarimetry}. A combination of polarimetry and with micro doppler signature (MDS), that reveals distinctive features of moving parts of objects have been used \cite{Kim2017ExperimentalSignature} to study drone detection by visually inspecting the returned spectrograms. MDS alone has been widely used for target sensing and classification. Visual inspection of MDS spectrograms in ground-to-air experiments have proven successful \cite{Rahman2018RadarW-band} for the detection of birds and drones on ranges up to 170m using K and W band radars, whereas other studies \cite{Molchanov2014ClassificationSignatures}, \cite{SvanteBjorklund2018TargetMicro-Doppler} use machine learning (ML) techniques for classification such as linear discriminant analysis and boosting, respectively, on shorter range experiments of up to 30m with X band radar configurations. More relevant deep learning approaches for data processing \cite{Rahman2020ClassificationImages}, \cite{Kim2017DroneImages} are also applied within a ground-to-air context on longer ranges of up to 100m and with radar configurations on the K band. An implementation of GoogLeNet, a CNN-based architecture has also been explored \cite{Rahman2020ClassificationImages} and applied to images and a series network to discretise between four classes: bird, drone, clutter and noise; using approximately 50K radar images. Results show an accuracy of around 99\% with LeNet, which slightly outperforms the series network. Other features such as a drone's distinctive MDS, both on the time and frequency domain, have been used to create neural network input representations and classify them with a CNN, obtaining high accuracy results \cite{Kim2017DroneImages}.
Radar cross section (RCS) is another fundamental radar feature that summarizes an object’s detectability by radar in the form of energy reflected back to the radar system. 

Radar representation, usually spectrograms, are generated by combining returned features such as range, azimuth, doppler and power. However, no end-to-end pipeline is in place, and sensing and classification is performed by visually exploring the returned spectrograms; typically on a ground-to-air setup. Drone sense and detect based on this type of data has been performed with radar configurations in the K band \cite{Shin2017ADetection}, \cite{RohlingUAVRadar}. However, the frequency modulated continuous wave (FMCW)  radar \cite{Shin2017ADetection} has a physical separation on both the receiver and transmitter and is hence not deployable in drones. Drone sense and detect at shorter ranges (approximately 30 to 90m) has also been explored with positive results \cite{Drozdowicz201635System}, utilizing two types of drones, where a wide doppler spread can be observed from the different moving parts which can then be used for target recognition and classification. Similar efforts and positive results have been reported \cite{TareqAl-NuaimLow-CostDetection}, \cite{BenjaminNussMIMODetection} from deployment of a multi-input multi-output radar (MIMO) operating at lower radar bands (S and C respectively). The small drones employed in these experiments can be detected at up to 100m and 30m respectively, proving the resilience of radar technology to a wide range of sensor setups.

Air-to-air use cases are much less explored, and typically limited only to tests within a few metres range, with results visually analysed. Studies on airborne radar sense and detect have focused on obstacles, including static features like walls and fences, and close range drone-to-drone sense and detect through visual inspection of spectrograms \cite{Moses2014UAV-borneAvoidance}. Drone-to-drone sense and detect has also been considered within a very short range (3 to 7m) and results were analysed by visually inspecting the returned spectrograms, and through ML algorithms such as linear discriminant analysis (LDA) \cite{Moses2011Radar-basedVehicles}. Recent studies \cite{Milias2023UAS-BorneApplications} have demonstrated radar sensing capabilities on longer ranges by detecting drones of two different sizes up to 500m using a custom X-band radar; however, no end-to-end detection and classification pipeline has been implemented.

Finally, some studies have recently started using DL approaches to radar data. Custom implementations of CNN-based neural networks have successfully been trained \cite{Roldan2020DopplerNet:Radar} on a dataset of 17K 11x61 distance vs doppler frequency images. The dataset is balanced to cover 3 classes: drones, people and cars. The results of the proposed CNN show improvements compared against two off-the-shelf networks (MobileNet and NasNet – both ImageNet trained) in all accuracy, precision and recall. The performance of a DL-based approach has also been compared with a ML counterpart \cite{Dale2020AnDrones}. The results suggest that a CNN can separate drones from non-drones to a higher degree of accuracy than a decision tree network, as it is capable of extracting deeper classification features. Finally, other studies have also examined deep learning on radar through the generation of synthetic data and the use of different deep learning techniques including CNNs \cite{Choi2018ClassificationSimulation}, temporal CNNs \cite{Brooks2018TemporalClassification}, and multi-layer perceptron (MLPs) \cite{Regev2017ClassificationNetwork} where the authors successfully classify targets (bird, single propeller UAV or a multi-propeller drone) and gain insight into drone parameters such as blade length; frequency of rotation; or how many blades and propellers the drone consists of. The main input source in these studies is radar data processed in RGB format to be used with DL architectures such as MLP, CNN and temporal modules. Other data input formats such as point clouds or point-based network architectures have not yet been explored as input data formats. Existing studies generally focus on classifying different isolated objects based on their characteristics without gaining scene understanding through multi object classification. 

Across all sensors, there is a need to build upon current efforts in target classification, sensing, and detection within aerial settings, progressing towards increased situational awareness. Our study extends the current state-of-the-art (SOTA) in radar, leveraging the technology’s capabilities by simultaneously classifying objects at mid/long ranges through the use of an end-to-end deep learning framework; and in applying the algorithm in more complex scenarios that require multi-object segmentation.

\subsection {Radar data processing and DNN architecture selection}

To date, most studies within the radar spectrum leverage a radar spectrogram to apply different processing techniques including ML and DL approaches that focus on MLPs and CNNs.

Despite offering a unique opportunity to address some of the challenges specific to the aerial context, point clouds have not yet been explored in an aerial context as a radar output format. Point clouds capture the 3D spatial distribution of radar returns, providing direct representations of the objects in the environment as individual points which translates into richer spatial information. Point clouds also constitute a more efficient data representation: they can be more compact than spectrograms, as they directly encode the spatial information of detected points. This can lead to reduced data storage requirements and potentially faster processing times.

However, point clouds' spatial structure in spherical coordinates comes with an additional challenge to shape them into fixed-size tensors for neural network training and inference. Typically, three paradigms can be used to treat 3D point cloud data:

\begin{enumerate}
    \item Projection-based: the 3D point cloud is projected into one of the 3 planes to create a 2D fixed-size image.
    \item Discretization-based: the extents of the point cloud are used to define a fixed size discrete representation, such as volumetric lattices, with voxelization being one of the most common. 
    \item Point-based: these methods act directly on the raw point clouds using a set of functions including shared multi-layer perceptron (MLPs) and global features to extract features. A fixed size input is achieved by a predetermined number of points contained in each sample.
\end{enumerate}

The sparsity of radar's aerial point clouds is a key setback of projection-based methods, which is observed by the amount of empty pixels when projected into a dense 2D image, triggering unnecessary computation. When considering discretization-based methods that employ a 3D volume instead of a 2D projection, this challenge increases proportionally to the size of the third dimension added. In our context of 3D aerial data, sparsity becomes an even more acute problem, since when an object is sensed, it is usually not large or close enough to build up the return count quickly for rich returns over short periods of time. Only when the ground or bigger obstacles like buildings are within the sensor’s FoV does the point cloud build up quickly (Figure 5 - bottom). In general this leads to highly sparse radar outputs.

In point cloud processing, the use of sparse convolutions \cite{LiuSparse} and kernel point convolutions \cite{ThomasKPConv:Clouds} have shown good performance for pointcloud processing in terms of accuracy and processing time \cite{Li2021AClassification} for lidar acquired data. However, the effect that sparser point clouds, such as radar's, have in the performance is less studied.

Point-based methods possess a unique advantage due to their capacity to operate directly on the point cloud data. Among these methods, PointNet \cite{Qi2016PointNet:Segmentation} stands out as a competitive choice. PointNet has demonstrated versatility and efficacy in a variety of applications, including segmentation and classification tasks. Recognizing its potential, we have opted to develop our implementation of a PointNet-based network, further enhancing its performance through an architecture optimization search tailored for aerial use cases.

\subsection {Neural network acceleration and performance optimization}

In aerial applications, in addition to the challenges of restricted bandwidth, low-latency requirements, and concerns about network stability, privacy, and security, a critical aspect is the SWaP (size, weight and power) constraints of UAV operations. Development and deployment of hardware and AI-based solutions for UAVs seek to strike a balance between the available resources of the former and the performance of the latter. Highly variable specifications, from small consumer drones to large industrial or military drones open the door to a wide range of computing units of different nature that can be integrated to meet that balance. While the preference is generally to use a GPU as a processing unit, these are typically power-hungry devices, and a relevant research question is exploring an efficient alternative, for instance, in situations of low power budgets common in UAVs. Current studies are exploring options like FPGAs \cite{Zhang2022FitNN:Drones}, \cite{Chen2022AnDrones} and specialised hardware \cite{Rojas-Perez2021TowardsCamera} as computer alternatives.

In this work, we employ a GPU processing unit using the SYCL (Single-source C++ Heterogeneous Programming for OpenCL) standard. A SYCL-based solution provides us with a versatile programming model for efficient code that's compatible with and portable to various hardware architectures \cite{Goli2020TowardsSYCL}, from GPUs to CPUs and FPGAs, hence addressing the portability challenge between different computing platforms. Additionally, it enables us to optimize neural network operations for our PointNet model, like convolutions, fully connected layers, pooling and activations, while abstracting the underlying hardware details.

\section{OUR APPROACH}
\label{sec:oa}

In this section, we present the steps for data acquisition and sensor performance characterization required to develop and train a DL algorithm for aerial radar point cloud segmentation.

\subsection {Scenario Setup}

The data acquisition and testing scenario takes place in an airfield south of Glasgow (Figure 1). This unique setting enables tests in an unsegregated airspace in and around an airfield. It provides a suitable environment to simulate UAM operational conditions, where returns of different classes are captured in varied and challenging configurations. All experiments were carried out within the current legal restrictions and with certified pilots.

\begin{figure}[H]
  \begin{center}
  \includegraphics[width=3.5in]{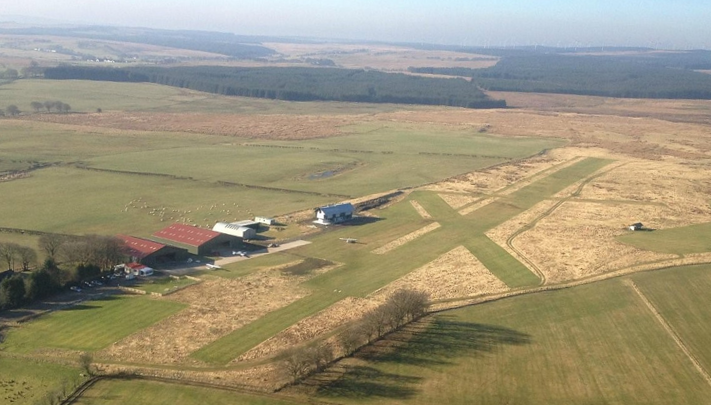}
  \caption{Scenario Setup - Strathaven Airfield. \textit{Image Source: Google Maps}}
  \label{fig:sim_opt_eff}
  \end{center}
\end{figure}

The airfield is equipped with a primary runway and two secondary runways. It encompasses typical airfield infrastructure, including buildings, weather instruments, and other objects that fall within the radar's detection range capabilities. Additionally, the radar captures ground returns, providing comprehensive observations of the airfield environment. An area for safe drone flying is designated next to the main runway which allows the safe flight of a larger M300 drone and a smaller recreational DJI mini. Airplane returns are captured taking advantage of airplanes on circuit in the airfield. A summary of the aerial targets considered in this study can be seen in Table \ref{tab:example}.

\begin{table}[h]
    \centering
    \caption{Aerial targets breakdown}
    \begin{tabularx}{\columnwidth}{|m{0.3\linewidth}|X|}
        \hline
        \includegraphics[width=\linewidth]{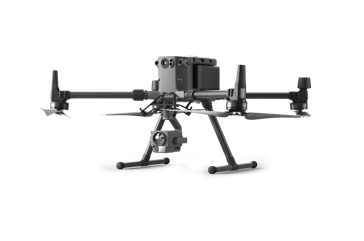} & 
        \textbf{DJI M300 RTK drone}
        \begin{itemize}
            \item Dimensions: 810 x 670 x 430 mm
            \item Weight: 6.3 kg (with two batteries)
            \item Materials: PVC, Carbon fiber
            \item Max speed: 23 m/s (83 km/h)
        \end{itemize} \\
        \hline
        \includegraphics[width=\linewidth]{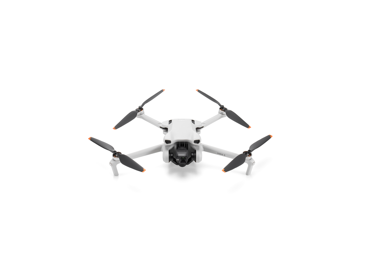} & 
        \textbf{DJI Mini drone}
        \begin{itemize}
            \item Dimensions: 245 x 289 x 56 mm
            \item Weight: 246 g
            \item Materials: PVC, Carbon fiber
            \item Max speed: 16 m/s (58 km/h)
        \end{itemize} \\
        \hline
        \includegraphics[width=\linewidth]{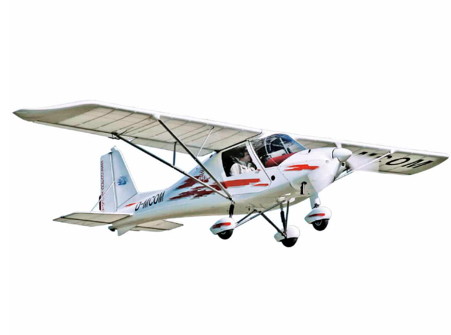} & 
        \textbf{Ikarus C42 aircraft}
        \begin{itemize}
            \item Dimensions: 6.2 x 9.45 x 2.34 m
            \item Empty weight: 265 kg
            \item Materials: Aluminum, composites
            \item Max speed: 175 km/h
        \end{itemize} \\
        \hline
    \end{tabularx}
    \label{tab:example}
\end{table}

\subsection {System Setup.}

In this novel air use case scenario, the entire system is developed from the ground up, encompassing every aspect from hardware sourcing to model deployment. The design process of the components takes careful consideration of the size, weight, and power (SWaP) constraints, which are particularly stringent in aerial systems. The final system on the radar-equipped drone consists of:

\begin{enumerate}
    \item Oculii EAGLE, a 4D imaging/multiple-input multiple-output (MIMO), 76-81GHz W-band radar. The system is conceived for an automotive use case, and applied to an aerial use case in this study. The antenna array arrangement yields a theoretical FoV that is considerably larger in the horizontal plane (up to approximately 120 degrees) than the vertical (approximately 25 degrees). It carries out the low-level signal processing internally and outputs a 3D point cloud. The choice was made amongst other models available as a good trade-off between sensing capabilities (range, azimuth, elevation and resolution) and suitable SWaP parameters for an air-to-air use case. Mechanical integration is achieved by designing and 3D printing a bracket that securely attaches to the host drone connections.
    \item Nvidia NX Xavier used for onboard processing and inference. At 70mm x 45mm size and 180g weight, it is within the allowable SWaP paramters. It also delivers up to 21 TOPS, making it a good candidate for high-performance computation of AI algorithms in embedded and edge systems.
    \item TP-Link antenna to send output to the ground controller. While data processing and inference are done onboard the drone, a ground laptop plus a WIFI access point to receive output from the model are used to send commands during data acquisition routines.
\end{enumerate}

\begin{figure}[H]
  \begin{center}
  \includegraphics[width=3.5in]{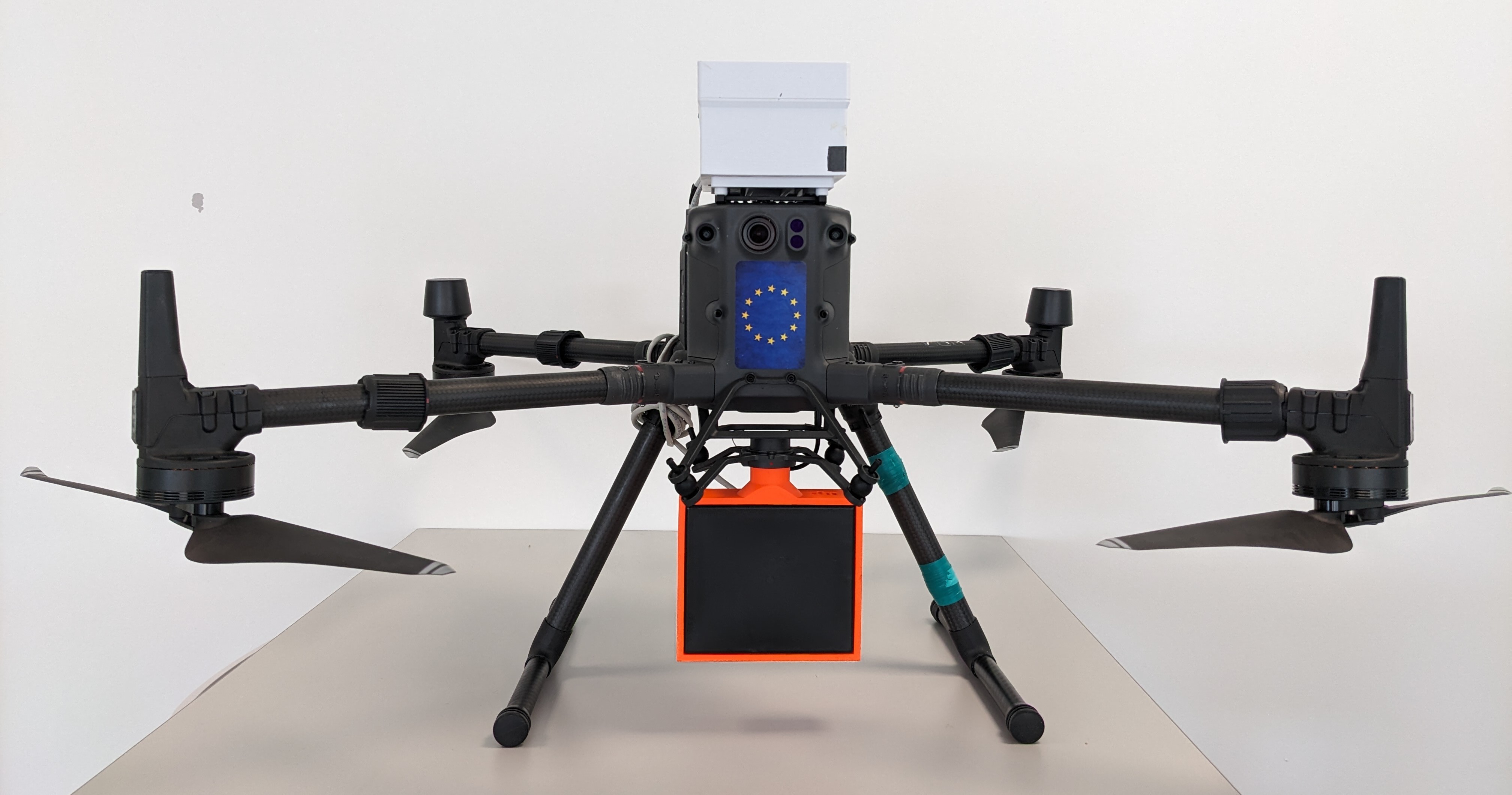}
  \caption{System setup - Radar-equipped drone. DJI M300 RTK mounted with white waterproof box (top) containing onboard computer and connections, and radar system (bottom) with custom orange attachment bracket.}
  \label{fig:sim_opt_eff_2}
  \end{center}
\end{figure}

Both the radar-equipped drone and one of the target drones (the DJI M300 RTK) are equipped with GPS capabilities which allow us to track their position over time. In addition, real-time kinematic (RTK) correction is used to improve the positioning accuracy of these systems.

The output of the sensor is a processed point cloud, hence we lack control over the raw returned signal. Therefore subsequent sections focus on measuring the performance of a point-based model on the data made available after internal sensor processing.

\subsection {Sense and detect performance characterization}

A performance characterization of the sensor is required as a means of establishing a boundary for true/false negatives which would impact the accuracy and recall scores. In the Oculii Eagle, the point cloud generation is abstracted with no access to the raw time, phase, and amplitude data. The detection FOV is not the same as the FOV specification sensor's data sheet as this is target-dependent. The sensor characterization allows us to differentiate between the system shortcomings and the actual DNN performance.

The 4D imaging radar used in our experiments has a three-dimensional directivity pattern that discloses range, elevation, and azimuth information about the target, in addition to native radar’s radial velocity and radar cross section (RCS). This defines a detection coverage for each target, as seen in Figure 3a, that can be approximated by two elliptic bicones, where their major axes are in the azimuth plane and the minor in the elevation plane. From one vertex where the radar-equipped drone is theoretically located (magenta point), the FoV opens with a certain angle to reach its maximum detection offset, both in range and elevation, a certain distance from the sensor, and then closes towards the center of the FoV to define the maximum detection range. Designed for an automotive use case, the antenna array arrangement places an emphasis on the azimuthal direction which means that the horizontal FoV area is considerably larger than the vertical, and the area with the strongest returns is coplanar to the radar’s orientation, degrading as the target moves away from it. 

The parameters that define a particular target detection coverage depend on characteristics such as the target’s size, shape and composition. Larger objects with larger RCS are easier to detect. RCS is also higher on targets with flat surfaces or edges perpendicular to the radar beam, as well as those made of metallic materials, as opposed to non-metallic materials like wood or plastic that may absorb or scatter radar waves. 

A first step to define the detection coverages for our controlled targets is required prior to the dataset acquisition exercises. This will inform the areas of interest that the target can fly with respect to the radar-equipped drone. It will also allow the characterization of the sensing and detection performance of the radar system and how it varies between targets of different characteristics.   

The detection coverage is approximated by running waypoint missions in the azimuth and elevation plane, as seen in Figure 3b. The target drones fly within these planes while the radar-equipped drone records the returns (Figure 3c). Defining the detection areas in these planes allows us to approximate the 3D volume where the target can be seen. The extent of the planes to be covered is defined by exceeding the theoretical FoVs provided by the radar’s datasheet for similar targets to ensure returns from all detectable areas are captured. Flying at high altitudes in the absence of other objects ensures that all returns captured belong to the target being studied. The granularity of the missions vary depending on the size of the target; from 1m between passes for the smaller DJI Mini, to 2.5m for the larger DJI M300. Coverage areas go up to 120m in range and 60m in offset.

The results of the experiment are summarized in Figure \ref{fig:perf_char}. Three pairs of plots are presented, one with the detection areas in azimuth and elevation for both targets, and rose plots with the return distribution across different radiation angles also in the azimuth and elevation detection planes. The M300 drone can be sensed up to almost 90m and 60 degrees horizontally at short ranges, whereas for the Mini, the maximum detection range is below 40m and 30 degrees horizontally at short ranges. In elevation, the area is considerably narrower, with the M300 drone being sensed at 20 degrees at short ranges and the Mini typically only when flying coplanar with the orientation of the radar. 

The RCS distributions provide differentiation in the characteristics of the target. In the case of the M300, the return distribution tends to cluster towards the higher end of the RCS values. On the other hand, for the DJI mini, the returns are comparatively weak, with the values concentrating towards the central and lower range of the RCS values. This trend holds true for all orientations and aligns with the higher reflectivity pattern expected for larger-size targets.


\begin{figure}
  \centering
  \subfloat[Fig. 3a]{\includegraphics[width=3.5in]{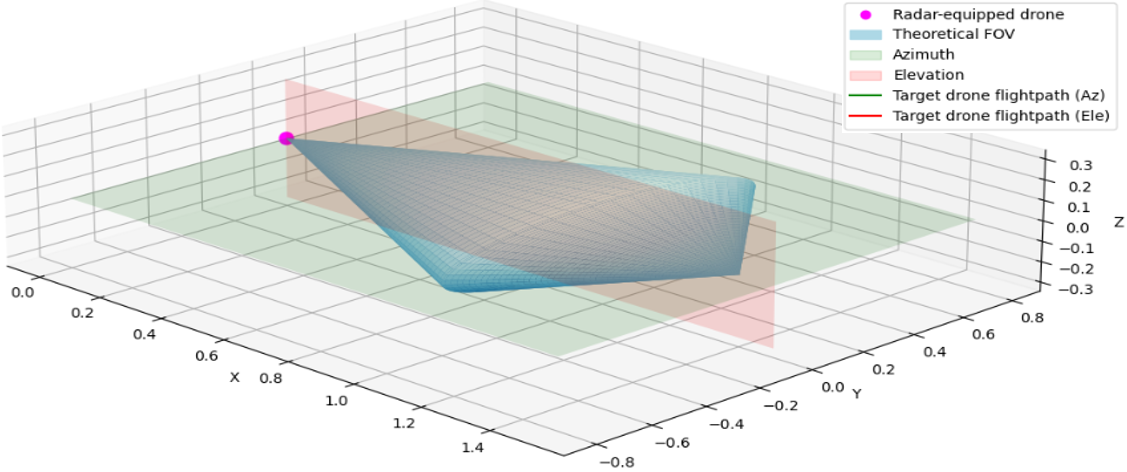}}\quad
  \subfloat[Fig. 3b]{\includegraphics[width=3.5in]{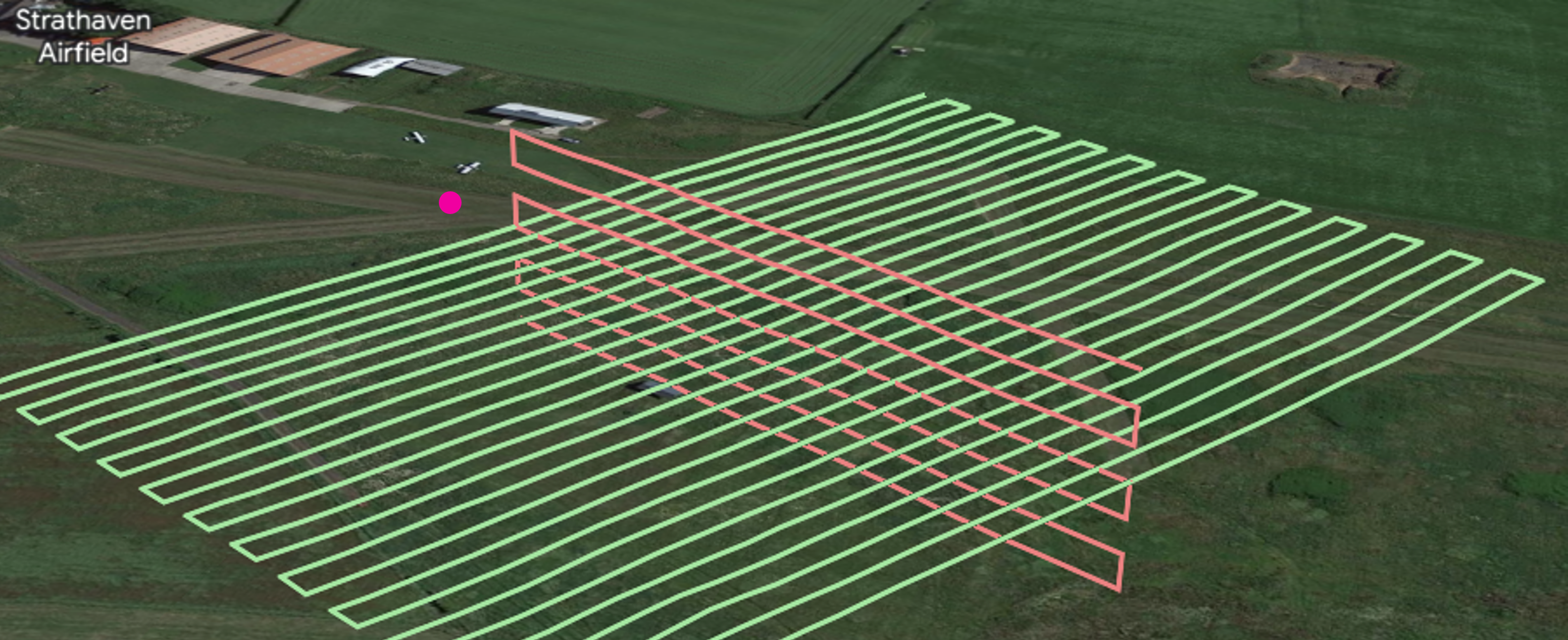}}\quad
  \subfloat[Fig. 3c]{\includegraphics[width=3.5in]{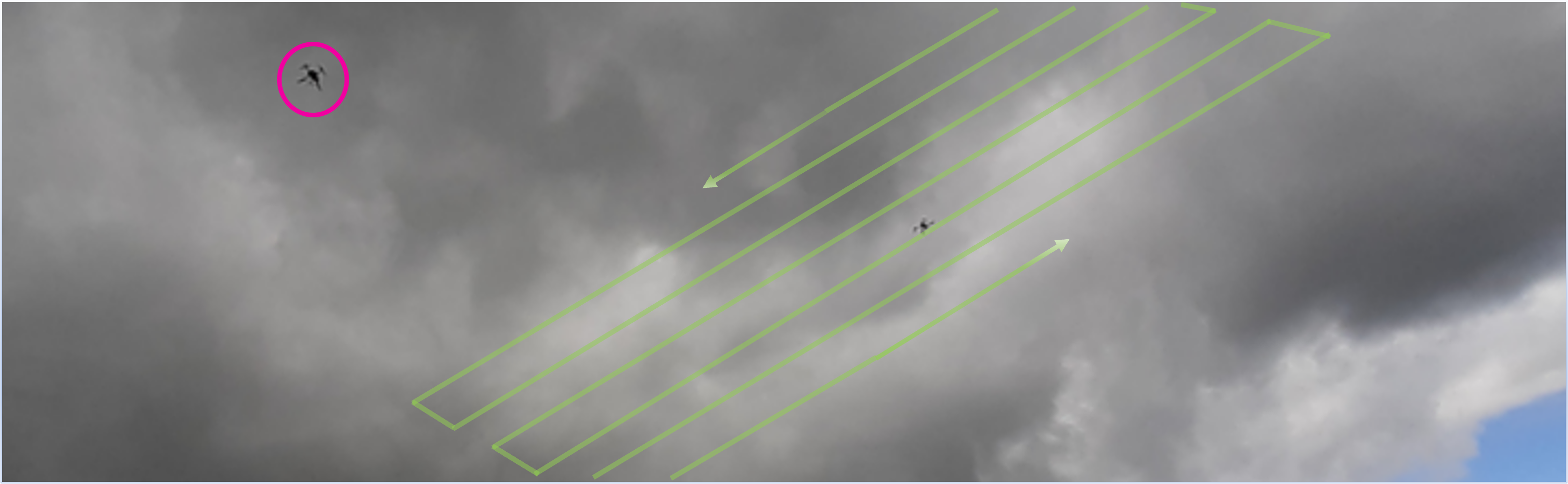}}
  \caption{Performance characterization steps of radar and controlled targets. 3a: theoretical detection coverage of radar defined by orthogonal planes to characterize the sensing volume. 3b: waypoint missions to cover the extent of the planes. 3c: execution of the missions.}\label{sim_opt_eff}
\end{figure}

\subsection {Data acquisition}

\begin{figure*}[t]
  \centering
  \includegraphics[width=\textwidth]{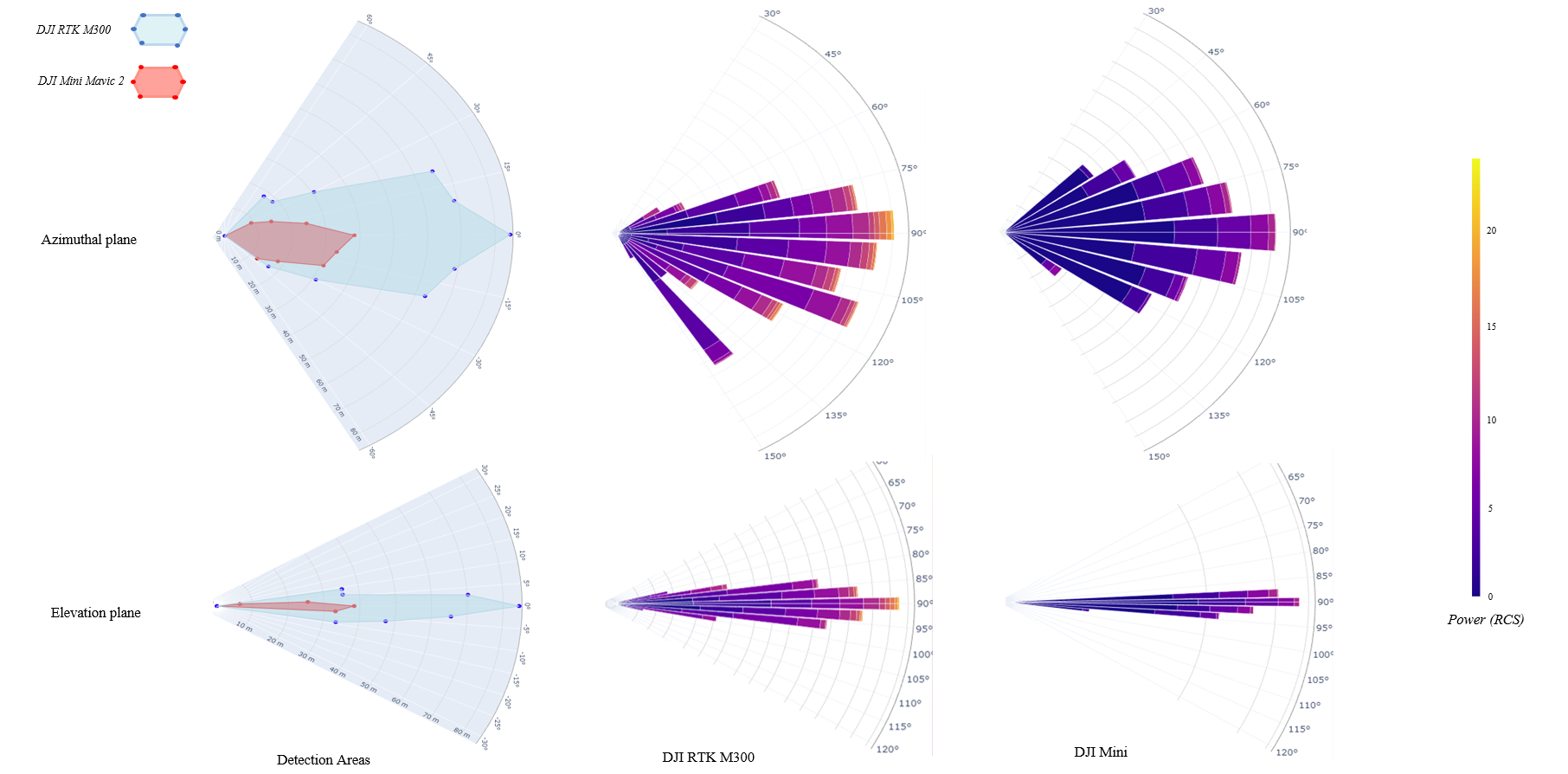}
  \caption{Performance characterization in the form of detection areas (first column) and power distribution (rose plot) for the larger DJI M300 (second column) and the smaller DJI Mini (third column). Top: performance in the horizontal plane/azimuth. Bottom: performance in the vertical plane/elevation.}
  \label{fig:perf_char}
\end{figure*}

To capture a rich dataset, pre-planned waypoint missions are used as well as manual flight manoeuvres with no repetitive patterns. In waypoint missions, the controlled targets fly through a specified flight path. The waypoint missions were developed through the volume of airspace approximated in Figure \ref{fig:perf_char} for each target, as well as incursion or exit manoeuvres into and out of it to acquire different sensing perspectives. Pre-planned missions included repetitive zigzag and radial patterns at varying ranges, angles and altitudes. This enables the accumulation of a varied set of returns and increased richness of the dataset.

Airplane returns are captured by rotating the radar-equipped drone towards the runaway to collect observations of planes on the circuit. Over the several passes captured in the dataset, the airplane performs the circuit stages of approach leg, landing / touch and go, and take off leg with slightly different trajectories which contribute variety to the data. The richness of perspectives is higher in the UAV-to-UAV observations, where the radar-equipped drone is orientated to maintain the observation, while the target drone approaches from several angles and with a certain degree of variation in the flight profile.

Ground returns as well as airfield infrastructure returns are recorded at times when the radar-equipped drone is flying closer to the ground, oriented towards buildings, or pointing towards airfield furniture located within its FoV. Mixed returns within the same beam scan and timestamp are common; making the future multi-object segmentation task more challenging, and with higher generalization capabilities.

\subsection {Dataset characterisation}
 The radar's recorded return count varies depending on the observed scene. At higher altitudes, minimal or no returns are typically observed. When a target, such as an approaching drone, is present, the returns build up to several tens per second. Conversely, when flying at lower altitudes and the radar waves interact with the ground or buildings, the return count rapidly increases to the thousand per second. This effect is depicted in the lower part of Figure 5.

The dataset contains a total of almost 1.2M radar returns which were acquired during approximately 2 hours of flight time. The dataset is highly imbalanced, reflecting the nature of the problem. The returns are split as shown in Table \ref{tab:ret_class}.

\begin{table}[ht]
\centering
\caption{Dataset Metrics Summary}
\begin{tabular}{|l|c|c|c|}
\hline
& Class & Number of returns & \multicolumn{1}{c|}{\%} \\
\hline
Ground & 1 & 779760 & 70.16 \\
\hline
DJI M300 RTK & 2 & 152065 & 13.68 \\
\hline
Airplane & 3 & 1487 & 0.14 \\
\hline
DJI Mini & 4 & 17539 & 1.58 \\
\hline
Infrastructure & 5 & 160589 & 14.45 \\
\hline
Total &  & 1111440 & 100 \\
\hline
\end{tabular}
\label{tab:ret_class}
\end{table}

Each radar return includes the target’s location in space in polar coordinates (range, azimuth, elevation), from where the cartesian (x, y, z) can be derived. The doppler signature, which corresponds to the radial component of the velocity vector, and the radar cross-section are also recorded. As the latitude longitude and altitude from the radar-equipped drone and the target drone are known, the object’s position can also be reconstructed in absolute coordinates for visualization of the results. 

A highly imbalanced dataset is obtained due to the varying complexity of building up the return count depending on the class. It also reflects the true nature of the problem at hand. Airplane returns are harder to obtain and therefore under and over-sampling methods have been considered to reduce the imbalance of this class. Oversampling the minority class from approximately 0.2\% to 2\% of the overall point count is shown to provide a good balance between performance and simplicity.

Data normalization plays a critical role in enhancing model performance and ensuring stable training. One important technique in this regard is feature scaling, which preserves the reference around 0 values, thereby maintaining the relative positions within the radar's field of view (FoV) and distinguishing between static and moving objects (zero and non-zero Doppler).

Finally, samples are split 75\% - 25\% into train and validation tests ensuring that a representative amount of the minority class is present in both splits.

\subsection {Labeling}

A per-point (return) labelling strategy is pursued to enable point cloud semantic segmentation. Different strategies for labelling the scenes are followed depending on the scene configuration and the targets involved. For targets where high frequency and accuracy GPS feed is available, an algorithmic strategy can be pursued for automatic labelling. In this case, it is applied to the DJI M300 RTK as follows.
Given two high frequency synchronized streams of GPS data: one for the sensor-equipped drone (denoted as \(D_s\)) and one for the target drone (denoted as \(D_t\)) of dimensions (w, l, h). 

For each synchronized timestamp \( t_i \), the global positioning and orientation of the sensor-equipped drone and the positioning of the target drone are defined respectively by:
\[
D_{s,i, \text{globl}} = (LAT_{s,i}, LON_{s,i}, ALT_{s,i}, \psi_{s,i})
\]
and
\[
D_{t,i, \text{glob}} = (LAT_{t,i}, LON_{t,i}, ALT_{t,i}).
\]

With a known ground station point \( St_{\text{lat,lon,alt}} \), the problem can be converted from a global to a local coordinate system. The new positioning becomes, respectively:
\[
D_{s,i, \text{loc}} = (x_{s,i}, y_{s,i}, z_{s,i}, \psi_{s,i})
\]
and
\[
D_{t,i, \text{loc}} = (x_{t,i}, y_{t,i}, z_{t,i}).
\]

Consider also a radar sensor that provides a high-frequency feed of time-synchronized radar returns relative to the sensor-equipped drone in local (polar) coordinates:
\[
RR_{i, \text{loc}} = (R_i, \alpha_i, \beta_i)
\]
with performance error parameters \( R_{\text{e}}, \alpha_{\text{e}}, \beta_{\text{e}} \). 

The task becomes to determine whether a given radar return \( RR_i \) should be labelled as belonging to the \( D_t \) class. The problem can be solved by measuring the distance between each return and the time-synchronized position of the target drone. If the distance is within an expected margin, the condition is met. Formally, the following inequality needs to hold true:

{
\small
\begin{align}
&\sqrt{({x_{rad,i} - x_{t,i}})^2 + ({y_{rad,i} - y_{t,i}})^2 + ({z_{rad,i} - z_{t,i}})^2} < \nonumber \\
&\sqrt{(\frac{l}{2} + R_{\text{ac}})^2 + (\frac{w}{2} + R_i \cdot \sin(\alpha_{\text{ac}}))^2 + (\frac{h}{2} + R_i \cdot \sin(\beta_{\text{ac}}))^2} + \epsilon
\end{align}

where
\begin{align*}
x_{rad,i} &= x_{s,i} + R \cdot \sin(\alpha_i + \psi_{s,i}) \cdot \cos(\beta_i),\\
y_{rad,i} &= y_{s,i} - R \cdot \sin(\beta_i),\\
z_{rad,i} &= z_{s,i} + R \cdot \sin(\alpha_i + \psi_{s,i}) \cdot \cos(\beta_i).
\end{align*}
}

On the left-hand side of (1), the inequality calculates the 3D euclidean distance between a pair of time-synchronized radar return-target drone positions. This distance needs to be within the maximum expected offset of a return belonging to the target being considered. This is calculated on the right-hand side of the inequality and accounts for the target size considered, as well as the known measurement errors of the sensor. In practice, additional sources of error can occur, including time shifts, positioning inaccuracies and movements of the radar-equipped drone which can affect a rigidly mounted system. These errors are accounted for in the \(\epsilon\) term. Depending on the setup configuration such as sensor characteristics or presence of IMU corrections \(\epsilon\) can take different values. In our study, we tested different values empirically and allowed for an additional 1.5m margin across all measurements.

This approach can be extended simultaneously to additional streams of GPS data such as other drones or planes equipped with GPS capability. Also, note that the algorithm assumes relatively stable operating conditions on the radar-equipped drone, with corrections only in the yaw axis. For a system operating in more varying conditions, corrections are likely required for pitch and roll as well.

 The labelling strategy can be visualized in Figure 5 where the target drone M300 is recorded performing radial movements over time (t0, t1). The blue crosses show the real GPS position of the target drone (M300). The dots represent radar returns; while the triangles represent the location of the radar-equipped drone. The colouring or labelling of the radar returns is done through the process described above. Radar returns with a matching time stamp equivalent to that of the target drone's GPS positioning, and within the allowable margin, are assigned 'DJI M300' class. Some scattered returns are picked up at ranges greater than 60m and labelled green (ground). At t2 the radar-equipped drone starts descending for landing, leaving the target drone outside of the FoV and therefore the points are labelled as ground.

Finally, airplane returns are labelled by identifying the returns contained within the area of airfield approach, landing, touch and go, and take-off manoeuvres, in absolute coordinates. A manual check is performed to make sure that no points are mislabelled.

All other scenes with the DJI Mini, ground and airfield infrastructure returns are labelled manually using video feed of the scene to reconstruct the position of each object and match it with that observed in the corresponding point cloud.

\begin{figure}
  \begin{center}
  \includegraphics[width=3.5in]{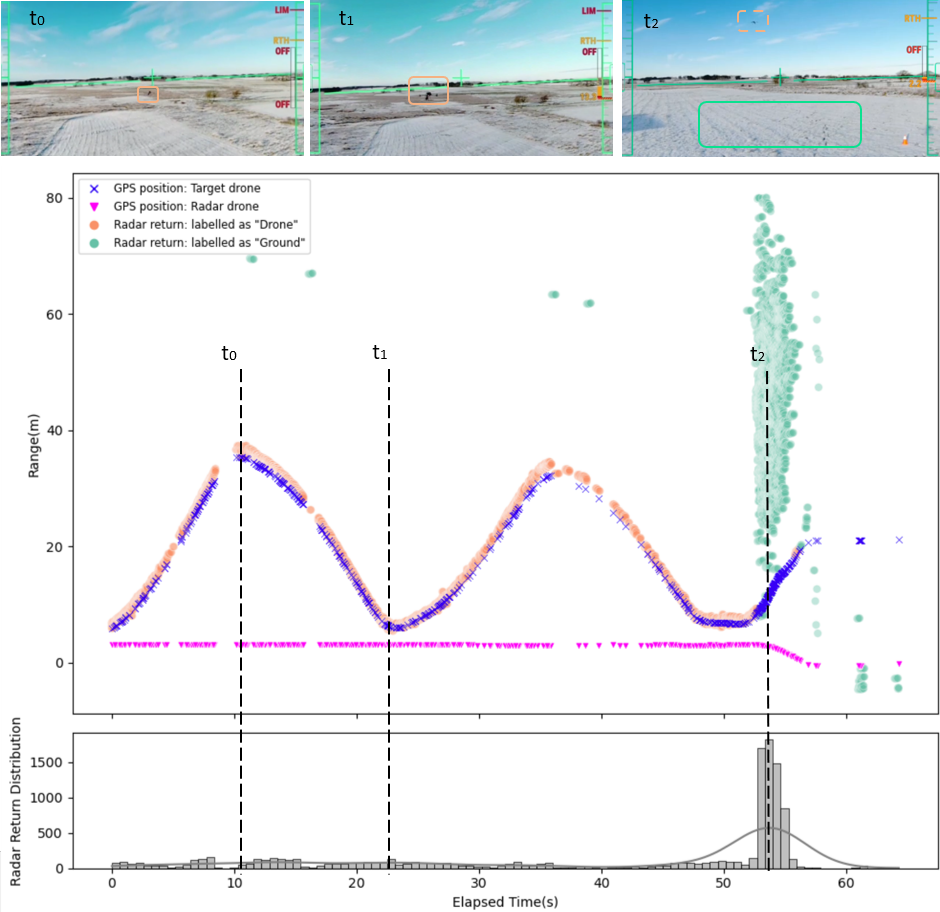}
  \caption{Self-labelling algorithm and return distribution over time. Pictures on top from radar’s drone first person view (FPV) camera show target drone flying at a longer range (t0), at a shorter range (t1), and landing (t2)}\label{sim_opt_eff}
  \end{center}
\end{figure}

\subsection {Input Representation for PointNet Learning}

During training, PointNet accepts a fixed-size input tensor which is achieved by pre-defining the number of points per sample as a hyperparameter. In the radar context, this presents a challenge due to the highly variable data throughput in an aerial context. Three options have been considered:

\begin{enumerate}
    \item Fixed timeframe with padded or sampled points: timeframe ‘tf’ and point count ‘pc’ are defined as hyperparameters. 
    \item Fixed point count with variable timeframes: a point count is defined, and the timeframe to fill it is variable.
    \item PointNet++'s \cite{Qi2017PointNet++:Space} KNN approach.
\end{enumerate}

Option three, which utilizes the PointNet++ KNN approach, is not preferable for our specific use. This method is primarily designed for down-sampling point clouds with high point counts and capturing local dependencies between neighboring points. However, aerial radar data often consists of sparser point clouds with well-separated classes. Additionally, returns of small or distant targets for a given timeframe are usually single-digit sets of points where the point count is a distinctive feature of the target itself, and applying KNN can lead to information loss, potentially hindering the network's ability to accurately classify such targets. Option two would produce long periods without updates, and be overburdened when the number of returns is high.

Option one is the preferred choice as it guarantees a fixed update rate. It comes with the challenge of dealing with samples over and under the desired point count within fixed timeframes. After studying the radar return feed, an approach shown in Algorithm 1 is proposed to prioritize the inclusion of returns with Doppler $\neq 0$ into the input tensor to capture all scarce information about the moving targets. The tensor is then completed by sampling or padding points to match a certain fixed point count. The specifics on how the data is packed for training and testing PointNet are provided in Algorithm \ref{alg:rpc_enc}.

\newcommand{\concat}{%
  \mathbin{{+}\mspace{-8mu}{+}}%
}

\begin{algorithm}
\footnotesize 
\caption{Aerial Radar Point Cloud Encoding for PointNet Training and Inference}
\label{alg:pointcloud}
\begin{algorithmic}[1]
\Require
  \State Point Cloud $P = \{p_i, p_{i+1}, \ldots, p_n\}$ 
  \State $p_i = (t_i, x_i, y_i, z_i, \sigma_i, v_i)$ \Comment{as $(time, x, y, z, power, velocity)$}
  \State timeframe = $tf$
  \State pointcount = $pc$
\Ensure $t_n - t_i = \text{tf}$
\Procedure{Process Point Cloud}{}

   $P_v \leftarrow \text{randomShuffle}(\{p \, | \, p \in P, \text{vel}(p) \neq 0\})$
   
   $P_{v\!'} \leftarrow \text{randomShuffle}(\{p \, | \, p \in P, \text{vel}(p) = 0\})$
   
   $Pr \leftarrow \text{take}(pc, \text{cycle}(P_v \concat P_{v\!'}))$

\State \textbf{return} $Pr$
\EndProcedure
\end{algorithmic}
\label{alg:rpc_enc}
\textbf{Algorithm 1 End}
\end{algorithm}

\subsection {PointNet Training}

A range of timeframes was tested and it was found that 200ms provided a good trade-off between a fast refresh rate, with sufficient point count accumulation for scene awareness. The dataset provides a total of 12K samples under this configuration. The number of points per sample is set at 256, striking a good balance between over and under-populated samples. The neural network is trained by shaping the data into a body frame perspective, passing each detection’s position in cartesian coordinates (x, y, z) respective to the host drone’s radar in addition to the RCS and the Doppler signature. The training batch size is set at 32, forming an input size for training of (32, 256, 5) per batch.

An initial learning rate of  1e-3 is chosen and \emph{Adam} is selected as the optimization algorithm \cite{Kingma2014Adam:Optimization}. The training of the network continues until the validation loss reaches a point of stability. 
At test time, the validation samples are passed through the network using the same approach as shown in Algorithm \ref{alg:rpc_enc} to obtain the performance metrics and ensure that the validation results seen during training are consistent in unseen scenes. 

\subsection {Network optimization and acceleration}
Finally, a significant challenge in the aerial realm is designing a compact hardware and software setup that meets the stringent SWaP (size, weight, and power) constraints, inherent to UAV operation, without compromising performance. 

In this work, we opt to equip our system with an NVidia GPU running with portDnn, an open-source, state-of-the-art SYCL (Single-source C++ Heterogeneous Programming for OpenCL)-based deep neural network (DNN) library designed to harness the computational power of heterogeneous computing platforms for accelerating DNN computations. Leveraging the SYCL standard, SYCL-DNN provides a high-level programming model that allows us to write portable and efficient code targeting a wide range of hardware architectures, including GPUs, CPUs, and FPGAs. Doing so, we gain control to optimize the implementations of various neural network operations, required by our PointNet architecture, such as convolutions, fully connected and pooling layers as well as activations. This allowed aligning the architecture with the unique needs of aerial use cases while abstracting the underlying hardware details. 

\section{RESULTS}
\label{sec:res}

In this section, we present the results of our approach, along with improvements considered for an aerial use case. We first compare our proposed system with a baseline model. Then we present an optimization of the network architecture and the performance of the model.

\subsection {Benchmark against ML baseline models}\

\begin{figure}
  \begin{center}
  \includegraphics[width=3.5in]{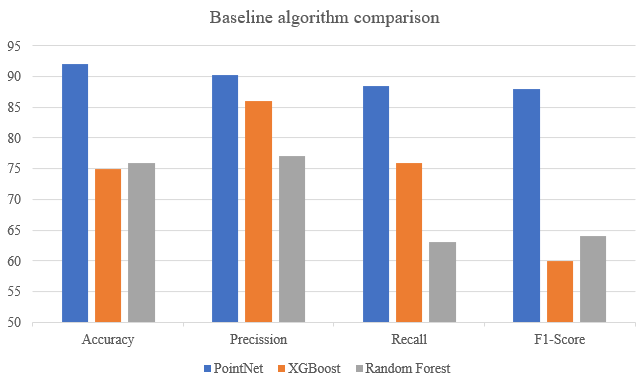}
  \caption{Performance comparison between machine learning and deep learning approaches.}
  \label{fig:ml_vs_dl}
  \end{center}
\end{figure}

In the context of aerial data, characterized by sparse and distinct classes, the effectiveness of simpler machine learning models compared to deep learning architectures comes into question. To investigate this, two multiclass classification models were evaluated using the curated train-test datasets. The first model, a decision tree-based algorithm - Random Forest, and the second, a gradient boosting framework - XGBoost, were chosen due to their distinct advantages. Both models possess the robustness to handle outliers effectively. Additionally, Random Forest naturally mitigates overfitting due to its ensemble nature, while XGBoost incorporates built-in regularization parameters to control model complexity, thereby curtailing overfitting. The performance is summarized in Figure \ref{fig:ml_vs_dl}.

\begin{figure}
  \begin{center}
  \includegraphics[width=3.5in]{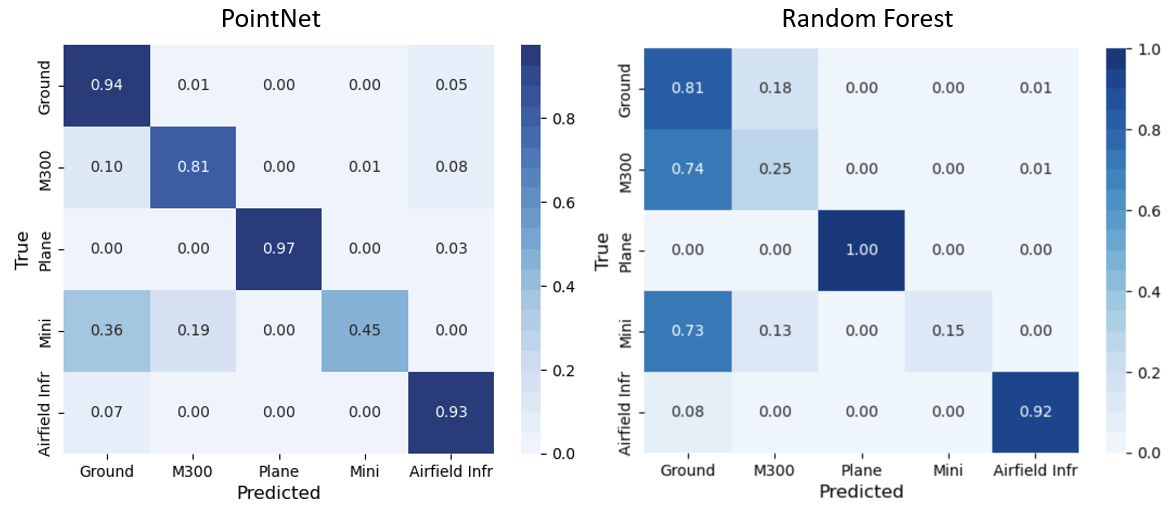}
  \caption{Confusion matrix comparison : ML(random forest) vs DL (Our PointNet implementation)}
  \label{fig:conf_mat}
  \end{center}
\end{figure}

Additionally, an implementation of PointNet largely mirroring the original paper is also included. As can be observed, our implementation of PointNet outperforms the ML models in all metrics, with an overall increase in accuracy of 15\%, hence justifying the design choice of a deep learning framework to tackle aerial point cloud segmentation. The failure modes of the ML learning algorithms can be observed by comparing the confusion matrices, as shown in Figure \ref{fig:conf_mat}.

PointNet is better able to capture the underlying features of the different objects across the range of scenes, perspectives and configurations of the dataset, As observed in Figure \ref{fig:conf_mat}, the random forest approach struggles greatly to correctly classify the two drones. In the case of PointNet, while the misclassification in the smaller drone are split between the larger drone and the class ‘Ground’, in the case of the ML approach, both drone classes are misclassified to a higher extent; with the ‘Ground’ class a more severe error. This may suggest that this approach is not fit to learn highly non-linear functions to capture the dynamics of an aerial use case.

\subsection {PointNet architecture optimization}\

An additional test investigated the suitability of our PointNet implementation network for aerial scenarios. Originally tailored for scenarios with closely positioned classes and object part segmentation in large-scale indoor scenes (ShapeNet \cite{Chang2015ShapeNet:Repository}), PointNet's architecture suits specific data characteristics. Transformation networks (TNets) in the model, including input and feature transforms, significantly contribute to the network's parameters, ensuring spatial alignment and enhancing orientation invariance at a scene and object level. While TNets excel in dense point cloud scenarios, sparse and less accurate radar data may benefit more from only scene-level transformations.
Furthermore, exploring the model's performance with reduced layer depth offers insights into training dynamics and effectiveness. To analyze the effects on training and performance, four models with varying complexity were defined. Results are summarized in Figure 8.

\begin{figure}
  \begin{center}
  \includegraphics[width=3.5in]{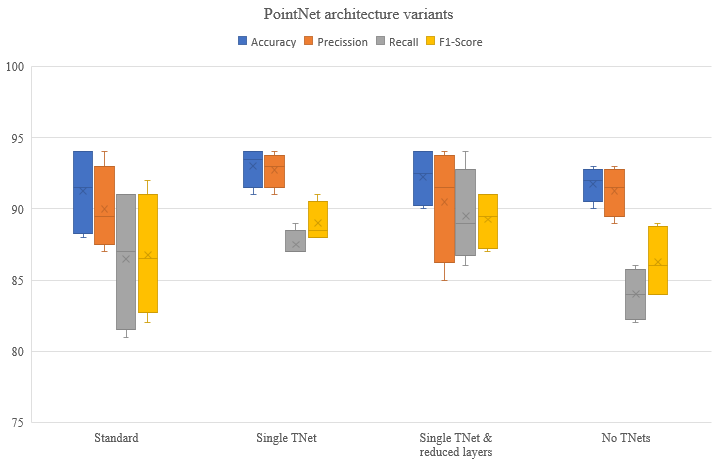}
  \caption{Performance comparison of different PointNet-modified implementations}\label{sim_opt_eff}
  \end{center}
\end{figure}

\begin{itemize}
    \item PointNet: retains the original implementation of the architecture. Number of params: 7.5M
    \item One TNet only: removes one TNet to reduce network complexity and capacity. Number of params: 2.3M
    \item One TNet and reduced layers: same architecture as above and halves the layers width. Number of params:1.25M
    \item No TNet: gets rid of both TNets. Number of params:1.5M
\end{itemize}

The boxplot shows that all four models achieve a reasonable performance, proving that networks that use MLPs as building blocks are a good option for aerial point cloud data. The addition of TNets also improves the performance, particularly across recall and F1-Score. The network with a single TNet yields the best overall performance, with the highest accuracy and precision, and average or above the average recall and F1-Score. It also has the most stable performance on the test set, with the lowest variability between the five different runs each setup was tested over. The chosen model classification performance parameters are shown below:

\begin{figure}
  \begin{center}
  \includegraphics[width=3.5in]{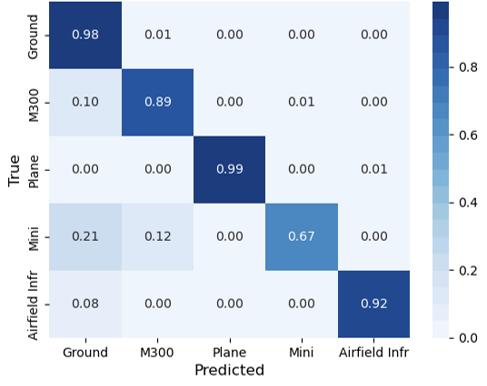}
  \caption{Confusion Matrix of optimized PointNet model}\label{sim_opt_eff}
  \end{center}
\end{figure}

The per-class parameters are presented in Table \ref{tab:pcp}.

\begin{table}[ht]
\centering
\caption{Per-class performance of the final model}
\begin{tabular}{|l|c|c|c|c|}
\hline
\cellcolor{gray!25} & \cellcolor{gray!25}Precision & \cellcolor{gray!25}Recall & \cellcolor{gray!25}F1-Score & \cellcolor{gray!25}Accuracy \\
\hline
\cellcolor{gray!25}Ground & 0.92 & 0.98 & 0.95 & 0.98 \\
\hline
\cellcolor{gray!25}DJI M300 RTK & 0.92 & 0.88 & 0.9 & 0.89 \\
\hline
\cellcolor{gray!25}Airplane & 0.98 & 0.99 & 0.98 & 0.99 \\
\hline
\cellcolor{gray!25}DJI mini & 0.86 & 0.67 & 0.75 & 0.67 \\
\hline
\cellcolor{gray!25}Airfield Infr & 0.99 & 0.92 & 0.96 & 0.92 \\
\hline
\cellcolor{gray!25}Combined & 0.93 & 0.89 & 0.91 & 0.94 \\
\hline
\end{tabular}
\label{tab:pcp}
\end{table}


\begin{figure*}[t]
    \begin{adjustwidth}{-\columnsep}{-\columnsep} 
        \begin{subfigure}{\textwidth} 
            \centering
            \includegraphics[width=\linewidth]{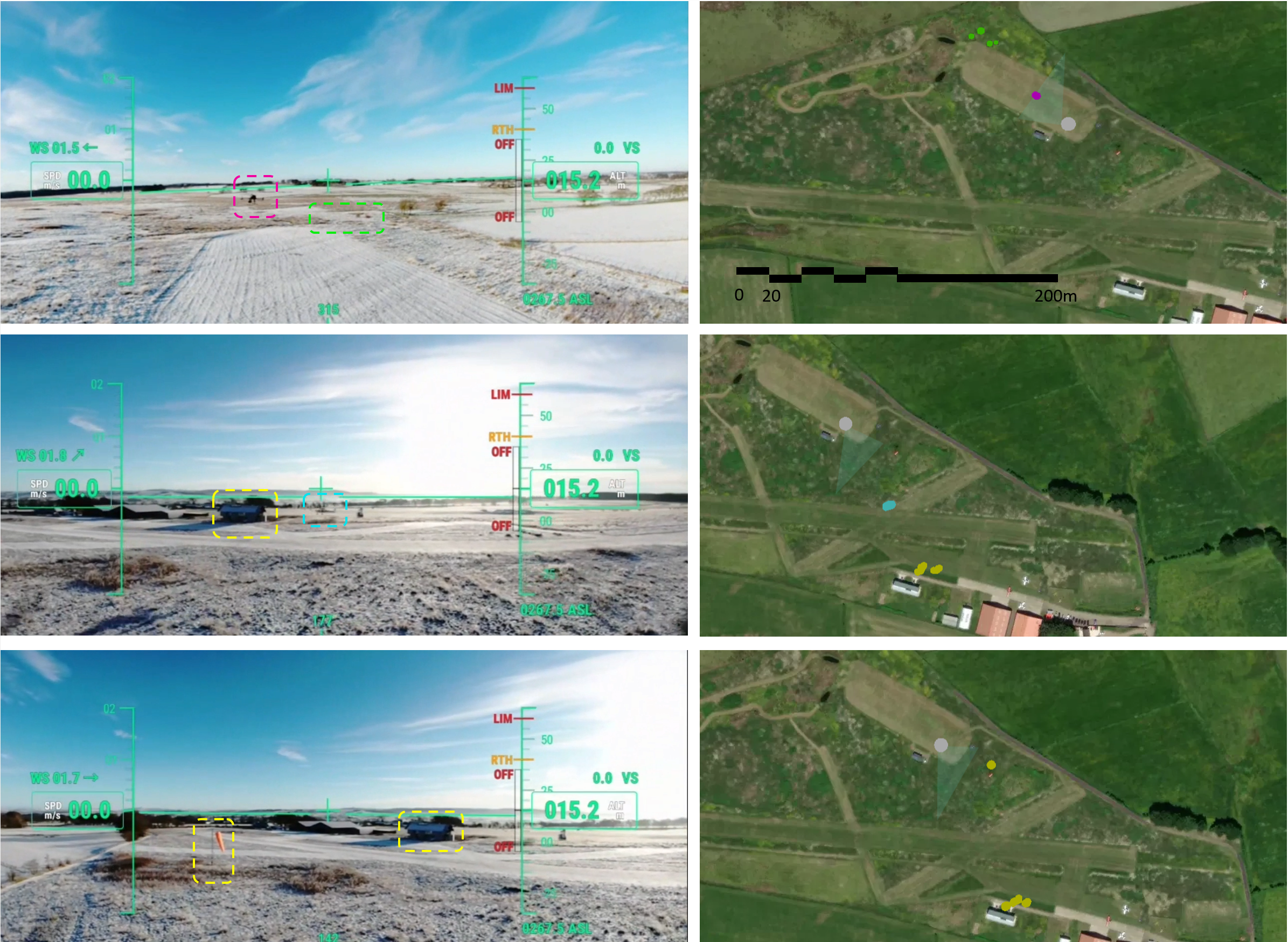}
        \end{subfigure}
        
        \begin{subfigure}{\textwidth} 
            \centering
            \includegraphics[width=\linewidth]{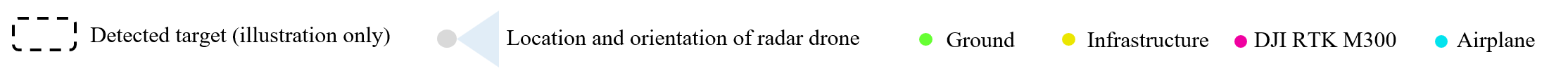} 
        \end{subfigure}
    \end{adjustwidth}
    \caption{Model predictions on left-out data projected in bird's eye view (BEV). Network results displayed on right hand side. Left hand side images are shown for illustration.}
    \label{custom_legend_label}
\end{figure*}

\begin{figure*}[t]
  \centering
  \includegraphics[width=\textwidth]{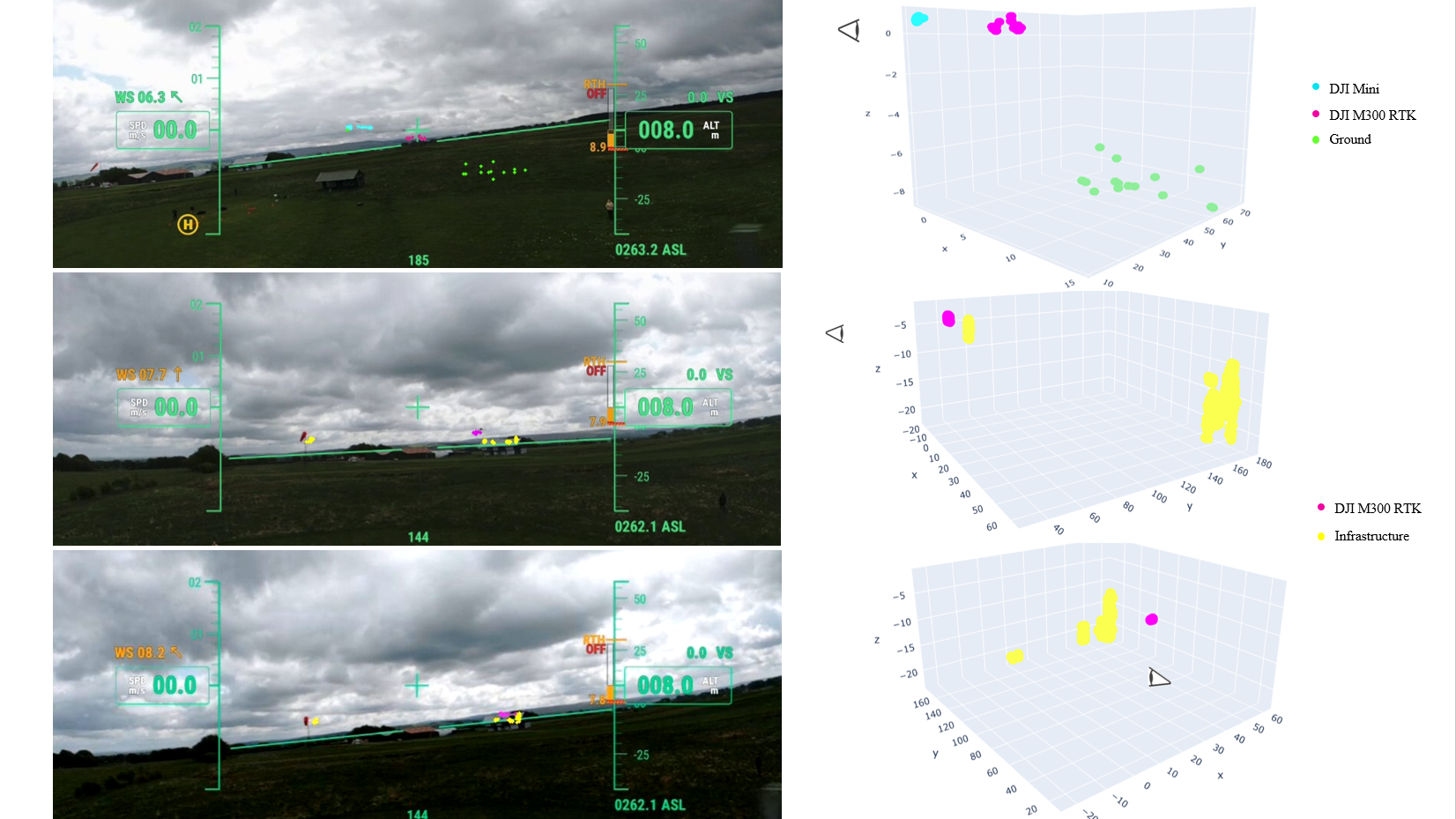}
  \caption{Results of the optimized PointNet architecture in the left-out data. Left, network predictions projected on the first person view (FPV), Right, 3D pointcloud visualized to get spatial information - viewer positioned in (0,0,0) looking towards positive y values.}\label{sim_opt_eff}
\end{figure*}

The findings depicted in Figure 9 and Table \ref{tab:pcp}` reveal a strong overall accuracy of 94\%, with consistently favourable metrics across all classes. The 'Mini' class has the most misclassifications, specifically with the larger drone class—expected due to its size and RCS returns—and the ground class, marking the most common misidentification.

In the context of aerial data, our PointNet implementation demonstrates promising results when applied to left-out data, as depicted in Figures 10 and 11. Figure 10 illustrates the projection of point clouds into aerial BeV (Bird's-eye View), to gain insights into the scale and ranges of the detections. As can be seen in the top figure, radar information is particularly useful for disclosing depth information about the targets (plane and building), which can be hard to discern from RGB data. It is then used to correctly classify the targets. Additionally, Figure 11 showcases the projection of points into the FPV, offering a pilot's perspective of how different targets are perceived in comparison to RGB data. These visualizations underscore the effectiveness and potential applicability of our PointNet-based approach in real-world scenarios, in particular, due to the camera's lack of depth information. 

\subsection {Neural network acceleration upgrades}

Leveraging portDNN, an open-source SYCL-based DNN library, we started from an un-optimized PointNet-SYCL model. Our efforts were centered around the operators that form the core of the PointNet architecture. This exploration highlighted opportunities for improvement that significantly enhanced the overall performance.

\begin{figure} 
  \begin{center}
  \includegraphics[width=3.5in]{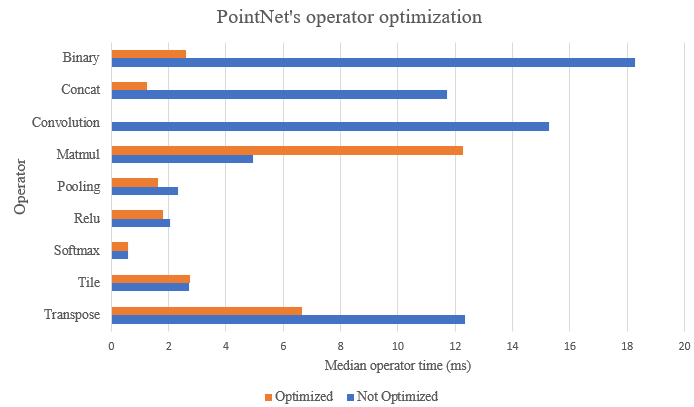}
  \caption{Performance comparison of optimized vs unoptimized PointNet implementations}\label{sim_opt_eff}
  \end{center}
\end{figure}

We directed our attention towards refining the implementation of operations such as Concatenation, Broadcasted Binary Operations, Softmax, and Pooling. Additionally, a notable improvement was realized by transforming 1D Convolutions into Matrix Multiplications, effectively reducing computation time. The improved performance can be seen in Figure 12 - note the convolution time reduced to zero in the optimized version.

\subsection {Model performance - Failure modes}

In this section we briefly describe some of the scene configurations that have been observed more likely to produce misclassifications. These can be used as a starting point for improvements in future work.

Figure 11 (top) shows a 'DJI M300 RTK' drone where some returns are misclassified as 'DJI Mini' drone. And a few returns from the 'DJI Mini' are misclassified as 'Ground' (visible in the image projection - (left)). The misclassification between a return classed as drone and 'Ground' or 'Infrastructure' has been observed to occur more often when the targets are static or moving at a slower speed. This shows that the velocity of the different targets acts as a key differentiator between classes. 

On the other hand, Figure 13 shows the predictions in three consecutive frames. Each frame constitutes 0.5s of accumulated data; while classifications are correct in the first and last frame, some misclassifications occur in the central frame, where a few returns are predicted as 'Ground' and 'DJI Mini'. 

\begin{figure} 
  \begin{center}
  \includegraphics[width=3.5in]{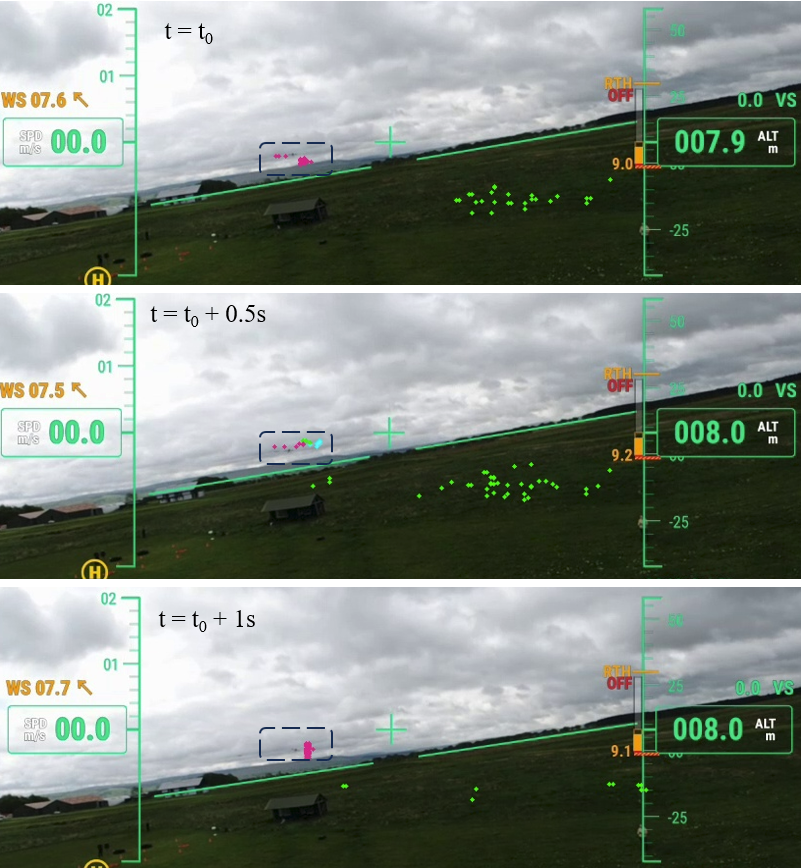}
  \caption{Mistakes on the network classifications. Each pointcloud represent 0.5s of data}\label{sim_opt_eff}
  \end{center}
\end{figure}

\subsection {Relevance of the sensing technology}

One final test in this study has been to incorporate the camera view alongside radar returns by projecting these together. This allows us to compare the perception of both sensing techniques. Camera offers a broader field of view, encompassing both horizontal and vertical FOVs. However, it has been observed to fall short of reliable sensing capabilities for small, rapidly moving targets, such as drones. Even with a relatively large resolution of (1024x768) commonly used in onboard systems, the camera has been observed to struggle to distinguish the DJI Mini at short distances (Figure 11, top). The situation is observed to be slightly better for larger targets like the M300 drone, identifiable at longer ranges. Nonetheless, moving at high speeds, both types of drones can close distances of tens of metres within a few seconds window, potentially leading to safety critical situations in an scenario where RGB technology alone is used. 

On the other hand, radar, though featuring a narrower field of view, has been observed to offer earlier warning capabilities. Additionally, radar returns tend to exhibit less noise compared to the dense pixel output from the camera. This characteristic makes them particularly suitable for interpretation by the point-based architecture presented in our study.

\section{CONCLUSIONS}
\label{sec:conc}

In this paper, we present an innovative application of airborne radar technology that advances UAVs ability to understand aerial scenes. To the best of our knowledge, this is the first effort to gain insights into evolving aerial scenes through multi-object point cloud segmentation of collision hazards.

Our main contribution is a customized PointNet-based neural network, enhanced with domain-specific knowledge for aerial contexts. Our implementation enables the classification of five unique classes in the aerial domain, with an overall segmentation accuracy of 94\%. This advances current state-of-the-art perceptual understanding capabilities, and enables UAVs to discern a range of objects; from mobile large drones (e.g., DJI M300 RTK), small drones (e.g., DJI Mini) and airplanes like the Ikarus C42 to static elements such as ground returns and man-made structures. This advancement allows UAVs to grasp scene configuration and dynamics effectively.

As part of this work, we introduce a workflow to evaluate the performance of the overall system. It includes a sensor characterization (4D imaging radar) and a novel methodology for aerial dataset acquisition and labelling of aerial point clouds. Through this characterisation and methodology, we are able to observe the varying performance of our system depending on the targets' characteristics. 

Finally, we carry out a network optimization to approach some of the limitations of deploying AI solutions within the aerial domain, where UAVs are characterized by low energy and computing budgets. To this end perform two optimizations at different levels of abstraction. At a higher level, we conduct a configuration search, aimining to strike a balance between building block types and sizes and overall performance. At a lower level, we carried out an operator optimization of our PointNet implementation. These optimisations yield significant gains in terms of performance. 

Overall, this study presents an end-to-end approach to multi-object segmentation of aerial point clouds focused in collision hazards. Our results pave the way for effective increased situational awareness, a requirement for unlocking BVLOS flights and deploying safe and efficient UAV systems in shared airspace.

\section*{Acknowledgments}

This work was funded by The Data Lab, Codeplay Software Ltd., and University of the West of Scotland under grant number Reg-201629 Intelligent RADAR Perception to Improve the Safety of Autonomous Vehicles (``The Data Lab Industrial Doctorate''), and by the European Commission under project RAPID (Risk-Aware Port Inspection Drones, 2020-2023) which is funded through the Horizon 2020 research and innovation programme and Grant Agreement number 861211.


%





\ifCLASSOPTIONcaptionsoff
  \newpage
\fi





\bibliographystyle{IEEEtran}
\bibliography{IEEEabrv,references}

\makeatletter
\def\@IEEEbiography[#1]#2{%
  \if@biography\else\@IEEEdblcol\fi
  \close@column@grid
  \ifIEEETOPSScover
    \vspace*{1ex}
    \else
  \vspace*{1ex}
    \fi
  \noindent\begin{minipage}{\textwidth}
    \IEEEPARstart{#1}{#2}
    \end{minipage}
  \@IEEEdblcol
  \vspace{0ex} 
  \fi}
\makeatother


\newpage
%

\vspace{-20pt} 
\begin{IEEEbiography}[{\includegraphics[width=1in,height=1.25in,clip,keepaspectratio]{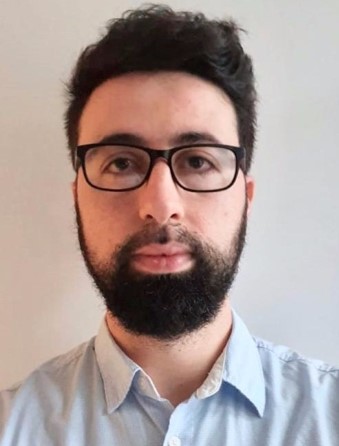}}]{Hector Arroyo}
received his B.Eng. and M.Eng. degrees in civil engineering from University of Burgos, Burgos, Spain and Edinburgh Napier University, Edinburgh, UK, in 2013 and 2014 respectively; and the M.S. degree in Data Science from University of Dundee, Dundee, UK, in 2021. He is currently working toward the Ph.D. degree in radar perception to improve the safety of autonomous vehicles with the ALMADA Drone Lab, University of the West of Scoltand and Codeplay, UK. His research interests include the study of sensing techniques for cross-modal, self-supervised learning applied to deep neural networks and robotics.
\end{IEEEbiography}
\vspace{-10pt} 
\begin{IEEEbiography}[{\includegraphics[width=1in,height=1.25in,clip,keepaspectratio]{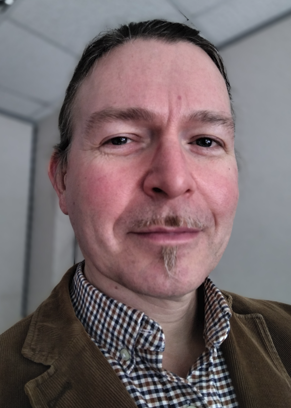}}]{Paul Keir}
received a B.Sc. (Hons.) degree in Instrumentation with Applied Physics from Glasgow Caledonian University, Scotland, UK, in 1995; an M.Sc. degree in 3D Computer Aided Graphical Technology Applications from Teesside University, England, UK, in 1996; an M.Sc. degree in High Performance Computing from the University of Edinburgh, Scotland, UK, in 2007; and a Ph.D. in Computing Science from the University of Glasgow, Scotland, UK, in 2012. He is currently a lecturer in Games Technology at the University of the West of Scotland, UK; with research interests in compilers, programming languages, and high performance computing.
\end{IEEEbiography}
\vspace{-5pt} 
\begin{IEEEbiography}[{\includegraphics[width=1in,height=1.25in,clip,keepaspectratio]{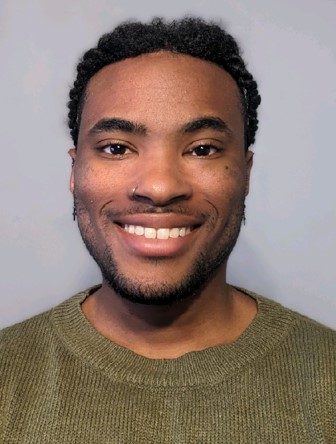}}]{Dylan Angus}
received a B.Sc degree with honours in Artificial intelligence and mathematics from the University of Edinburgh in 2019. It was this same year that they joined Codeplay focussing on accelerating sensor fusion and ML algorithms using SYCL.
During their time at Codeplay, Dylan has worked primarily on enabling and accelerating different DNN algorithms for frameworks such as oneDNN, GLOW and ONNXRuntime. Currently, their focus is on extending and optimizing Codeplay's DNN library written in pure SYCL, portDNN, and ensuring it is performant for various GPUs and accelerators.
\end{IEEEbiography}
\begin{IEEEbiography}[{\includegraphics[width=1in,height=1.25in,clip,keepaspectratio]{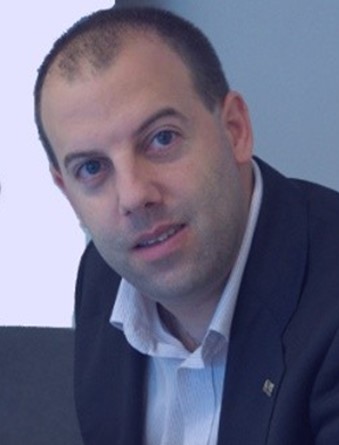}}]{Santiago Matalonga}
received his Engineer’s degree in Computer Science at the Universidad ORT Uruguay, Motevideo, Uruguay in 2003, his Ph.D. in software and systems from the Universidad Politécnica de Madrid, Madrid, Spain in 2011 and post-doctoral studies at COPPE/Universidad Federal do Rio de Janeiro, Brazil, in 2017. He is currently a Lecturer at the University of the West of Scotland and has over 70 publications in international journals and conferences. Dr Matalonga research interests include the management of the quality of Modern Software Systems; software architecture and enhancing the productivity of the development teams. He has strong industrial experience, some of his most recent engagements include coaching in software development lifecycle management, and lean-based standards adoption.
\end{IEEEbiography}
\begin{IEEEbiography}[{\includegraphics[width=1in,height=1.25in,clip,keepaspectratio]{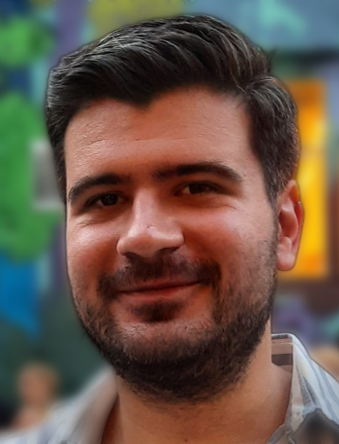}}]{Svetlozar Georgiev}
received the B.Eng. degree with Honors in Software Engineering from Edinburgh Napier University in 2019. In the same year, he embarked on his professional journey at Codeplay, where he worked on accelerating ML and DNN libraries and frameworks using SYCL. 
His current focus is on enhancing Codeplay's SYCL-based deep neural network library portDNN by implementing more DNN algorithms and optimising the performance of existing ones on different accelerators and GPUs.
\end{IEEEbiography}
\begin{IEEEbiography}[{\includegraphics[width=1in,height=1.25in,clip,keepaspectratio]{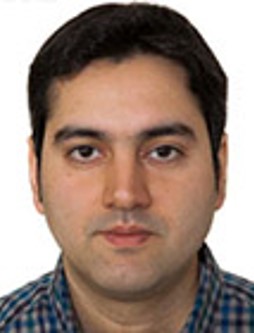}}]{Mehdi Goli}
received his B.Eng. and M.Eng. degrees in software engineering from Shahid Beheshti University, Tehran, Iran, in 2005 and 2009 respectively; and the M.S in computer science from The University of Edinburgh, Edinburgh, UK in 2011. He completed his Ph.D. at the Robert Gordon University, Aberdeen, UK, in 2015. He joined Codeplay in 2017 as a Senior Software Engineer in AI Parallelisation and he was the Team Lead of Eigen, SYCL-BLAS, and Nvidia backend for Intel oneMKL and oneDNN. Mehdi is currently the Vice president of research and development at Codeplay. He is responsible for leading impactful, influential, and innovative research and development projects, ensuring Codeplay remains a leading independent provider of AI and HPC enablement.
\end{IEEEbiography}
\begin{IEEEbiography}[{\includegraphics[width=1in,height=1.25in,clip,keepaspectratio]{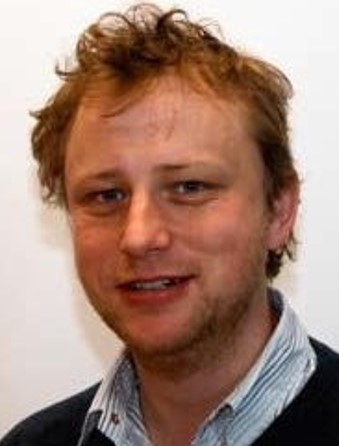}}]{Gerard Dooly}
received the B.Eng. degree in Electronic and Computer Engineering  and a Ph.D. working with optical based sensor systems at the University of Limerick, Limerick, Ireland in 2003 and 2008 respectively, He is a lecturer in digital technologies at UL and has worked extensively in robotics at UL for over 20 years. Dr. Dooly is Co-director in the CRIS Centre and is an Investigator in both the SFI centre for Smart Manufacturing, CONFIRM, and the SFI MaREI Centre for Energy, Climate and the Marine. His research includes real-time 3D reconstruction, machine vision and machine learning, optical sensors, structural health monitoring and automated survey and intervention. He is focused on the design and development of robotics and has led numerous mulit-partner, multi-agency projects both here in Ireland and on the continent.
\end{IEEEbiography}
\begin{IEEEbiography}[{\includegraphics[width=1in,height=1.25in,clip,keepaspectratio]{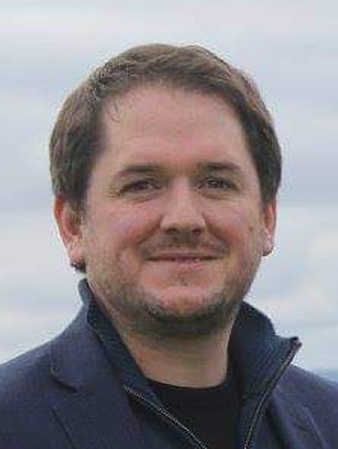}}]{James Riordan}
received the B.Eng. degree in Electronic Engineering in 2002, and the PhD. degree in underwater sonar modelling and simulation in 2006 at University of Limerick, Limerick, Ireland. He is currently a Reader at University of the West of Scotland where he leads the ALMADA Drone Lab. He is Principal Investigator of the European Commission H2020 project RAPID (risk aware port inspection drones) and The Data Lab Industrial Doctorate ``RADAR Perception to Improve the Safety of Autonomous Vehicles''. His research is primarily focused on advancing artificial intelligence techniques to improve sensor perception in drones and has applications in fields such as environmental monitoring, search and rescue, agriculture, and infrastructure inspection.
\end{IEEEbiography}

\end{document}